\documentclass{article}

\PassOptionsToPackage{numbers,compress}{natbib}

 \usepackage[main, final]{neurips_2026}

\usepackage[utf8]{inputenc} %
\usepackage[T1]{fontenc}    %
\usepackage{hyperref}       %
\usepackage{url}            %
\usepackage{booktabs}       %
\usepackage{multirow}       %
\usepackage{makecell}       %
\usepackage{amsfonts}       %
\usepackage{amsmath}        %
\usepackage{amssymb}        %
\usepackage{nicefrac}       %
\usepackage{microtype}      %
\usepackage[table]{xcolor}  %
\usepackage{graphicx}       %
\usepackage{subcaption}     %

\newcommand{\todo}[1]{}

\usepackage{xspace}
\usepackage{amsthm}

\definecolor{HeaderBlue}{RGB}{233,242,255}
\definecolor{RowGray}{RGB}{246,246,246}
\definecolor{SOTARed}{RGB}{215,35,45}
\definecolor{PrevBlue}{RGB}{42,96,255}
\definecolor{ConfCyan}{RGB}{0,153,204}

\title{PickMoment: Continuous-Time Single-Image-to-Video \\via Learning Deblurring and Blur-to-Video}

\author{%
  Junseong Shin$^{1}$ \quad Hyeonsu Jo$^{2}$ \quad Daehyun Kim$^{3}$ \quad Tae Hyun Kim$^{1,3\dagger}$ \\[0.5em]
  $^{1}$Department of Artificial Intelligence, Hanyang University \\
  $^{2}$Department of Intelligence Convergence, Hanyang University \\
  $^{3}$Department of Computer Science, Hanyang University \\
  \texttt{\{junsung6140, gustn249, daehyun, taehyunkim\}@hanyang.ac.kr} \\[0.3em]
  {\small $^{\dagger}$Corresponding author}
}

\begin{document}

\maketitle

\begin{abstract}
Motion blur arises from the uniform temporal integration of a continuous sharp signal over a finite exposure window, yet existing learning-based methods sidestep this physical model and predict only the sharp signal itself: most single-image deblurring methods recover a single frame at the exposure center, while blur-to-video methods predict a fixed set of frames. 
We introduce PickMoment, a continuous-time reformulation that directly learns the interval-mean blur over arbitrary sub-intervals of the exposure with a single deterministic model. 
Drawing an analogy to MeanFlow's average-velocity formulation, we train the model with three supervisions derived from the blur integral: an empirical reconstruction loss from available subframes, an additivity loss that enforces self-consistency across overlapping sub-intervals, and a sharp-frame loss anchored at the zero-interval limit. 
A single trained model unifies single-image deblurring, blur-to-video generation, and continuous-time pick-a-moment recovery as different queries to the same network, with no separate training for each task. 
Our PickMoment achieves state-of-the-art performance among generative-prior-based deblurring methods on GoPro and HIDE while competitive against restoration-based methods on RealBlur, and the highest per-frame fidelity on GoPro-7 blur-to-video, all with one forward pass per output frame and no iterative sampling.
Our project page is available \href{https://junsung6140.github.io/pickmoment_project_page/}{here}.
\end{abstract}

\vspace{-0.5em}
\section{Introduction}
\label{sec:intro}
\vspace{-0.5em}
Motion blur is a pervasive degradation in real-world imaging, arising from camera shake or object motion during the exposure process. 
Beyond deteriorating visual quality, it hinders human perception and limits the applicability of images in downstream tasks such as object detection~\cite{sayed2021motionblur}, 3D reconstruction~\cite{ma2022deblurnerf}, and depth estimation~\cite{hu2014joint}.
To address this, prior works have not only advanced learning-based deblurring approaches~\cite{nah2017deep} but also sought to understand the underlying principles of blur formation~\cite{fergus2006removing}.

Motion blur arises from the temporal integration of sharp frames over a finite exposure window.
Following Nah et al.~\cite{nah2017deep}, given an exposure interval $[0,T]$ of duration $T$, we model the blurred image $B(0,T)$ as
\begin{equation}
    B(0, T) \;=\; \frac{1}{T}\int_0^T S(u)\,\mathrm{d}u \simeq \frac{1}{M} \sum_{i=0}^{M-1}S_i,
    \label{eq:blur_concept}
\end{equation}
where $S(u)$ denotes the sharp frame at instant $u \in [0,T]$ and $M$ is the number of uniformly sampled frames used in discretization.
This physically grounded interpretation has been widely adopted not only for realistic blur synthesis, but also as a key principle for learning-based deblurring.
Building on this view, single-image deblurring methods have progressed from CNNs~\cite{nah2017deep,tao2018scale} and transformers~\cite{zamir2022restormer,kong2023efficient} to diffusion-based approaches~\cite{chen2023hidiff,liu2025fidediff}, achieving strong per-pixel and perceptual fidelity.
However, they still recover only a single sharp frame from the sharp sequence $\{S_0, S_1, ..., S_{M-1}\}$ (typically the center frame of the exposure), collapsing the temporal integration into a single moment and discarding the rich temporal structure encoded in the blur. 

\begin{figure*}[t]
    \centering
    \includegraphics[
        width=\textwidth,
        clip
    ]{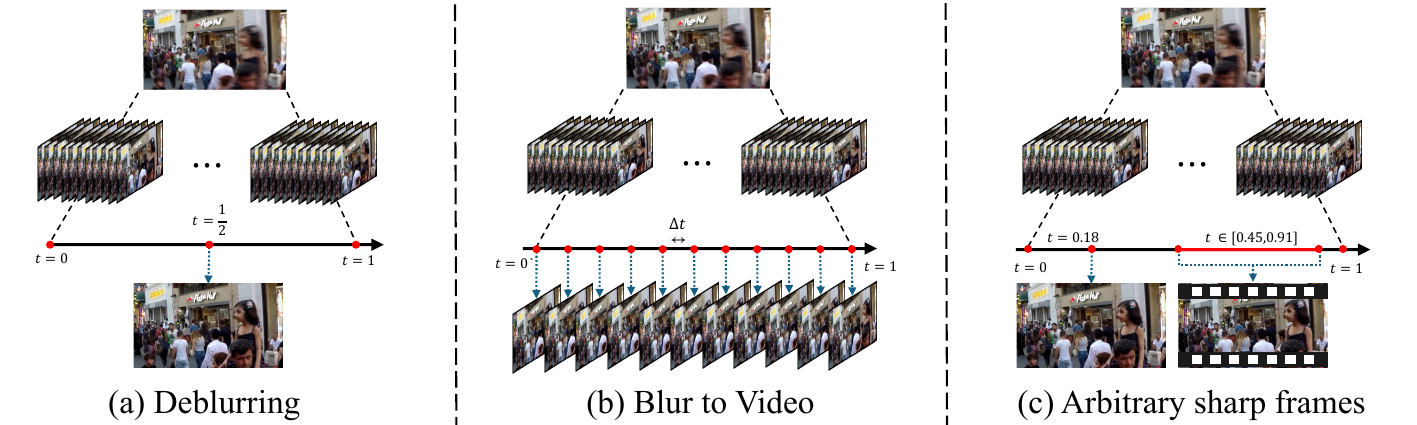}
    \caption{
    Overview of the proposed \textbf{PickMoment} framework.
    Given a single motion-blurred image, our method unifies three tasks:
    \textbf{(a)} Image deblurring, predicting the center sharp frame of the blur exposure.
    \textbf{(b)} Blur-to-video generation predicts an $n$-frame sharp video by uniformly sampling the normalized exposure trajectory at $M$ time points with interval $\Delta t = 1/(M-1)$; the figure illustrates an 11-frame example.
    \textbf{(c)} Arbitrary sharp-frame prediction enables the recovery of sharp frames at arbitrary time points within the blur exposure, thereby enabling continuous-time video generation. 
    }
    \vspace{-1em}
    \label{fig:overview}
\end{figure*}

To exploit this structure, blur-to-video methods aim to recover the entire sharp sequence from a single blurry image. 
Early work~\cite{jin2018learning} trains with order-invariant losses that accept both forward and reversed sequences as valid targets.
Subsequent methods introduce learned motion latents~\cite{purohit2019bringing}, differentiable warping with auxiliary regularizers~\cite{argaw2021restoration}, latent-space ordering classifiers~\cite{pham2023hypercut}, or, most recently, sampling from a pretrained video diffusion prior~\cite{tedla2025blur2vid}.
Across both single-image deblurring and blur-to-video tasks, however, the dominant learning target has remained the sharp signal itself, either at one timestep or at a fixed set of discrete timesteps.
This fails to fully exploit the temporal ordering encoded in the physical formulation of Eq.~\ref{eq:blur_concept}, where the model is used only for synthesizing training data rather than shaping the network predictions themselves.
In contrast, rather than learning a direct mapping from blur to a sharp frame or a sequence of video frames, our goal is to directly predict the blur formation integral itself from a single blurred image, allowing the physical model to serve as a core component of the learning objective.

A closely related idea has been explored in generative modeling. 
MeanFlow~\cite{geng2025meanflow} directly predicts an average velocity field supervised by identities derived from its integral definition over arbitrary intervals. 
Since motion blur, as defined in Eq.~\ref{eq:blur_concept}, is itself a continuous-time integral of sharp frames over the exposure interval, it shares the same algebraic structure as MeanFlow's average velocity field, such that integral-derived identities serve as natural self-consistency constraints across all sub-intervals.

In this work, we propose \textbf{PickMoment}, which trains a network to predict blur observations over arbitrary sub-intervals of the exposure duration, enabling the learning of continuous temporal representations that generalize to any temporal range from a single blurry input. 
Specifically, we leverage two structural properties of motion blur: the blurry input itself corresponds to the blur integrated over the full exposure interval, and standard datasets~\cite{nah2017deep, shen2019hide} provide subframes with known temporal ordering. 
Together, these properties, combined with our MeanFlow-inspired training strategy, enable the network to produce deterministic and direction-consistent reconstructions with one forward pass per output frame. 
As illustrated in Fig.~\ref{fig:overview}, single-image deblurring, blur-to-video generation, and arbitrary-moment recovery naturally emerge as different queries to the same model. 
Furthermore, sharp frames at arbitrary continuous timesteps arise as the zero-interval limit, enabling continuous-time \textbf{``pick-a-moment''} recovery from a single blurred image.

In summary, our contributions are as follows:
We introduce \textbf{PickMoment}, a physically grounded reformulation of motion deblurring that models a blurry image as a continuous-time integral over the exposure interval and casts deblurring as the prediction of interval-mean sharp content over arbitrary sub-intervals. 
By learning a continuous temporal representation, PickMoment unifies single-image deblurring, blur-to-video generation, and continuous-time \textbf{``pick-a-moment''} recovery within a single formulation, without requiring separate models or objectives for different tasks. 
To effectively learn this formulation, we draw a structural analogy between sub-interval blur prediction and the average velocity formulation of MeanFlow.
Specifically, we exploit two integral-derived identities --- additivity across sub-intervals and the zero-interval limit --- as supervision signals for temporally coherent learning. 
Extensive experiments demonstrate that PickMoment achieves state-of-the-art performance among generative-prior-based deblurring methods on GoPro and HIDE while remaining competitive with restoration-based baselines, and enables 1-step continuous-time blur-to-video generation with significantly reduced computational cost compared to video diffusion approaches.

\vspace{-1em}
\section{Related Work}
\label{sec:related}
\vspace{-0.5em}
\paragraph{Single-image motion deblurring.}
Single-image motion deblurring aims to recover a sharp image from a single blurry observation, evolving from traditional handcrafted priors~\cite{fergus2006removing,pan2016blind} to deep learning.
End-to-end CNN architectures~\cite{nah2017deep,tao2018scale,zhang2019deep,kupyn2019deblurgan,zamir2021mprnet,chen2022simple,fang2023ufpnet,mao2024adarevd} pioneered direct blur-to-sharp mapping, followed by transformer-based models~\cite{zamir2022restormer,tsai2022stripformer,kong2023efficient} that achieve state-of-the-art per-pixel fidelity through global self-attention.
More recently, diffusion-based approaches have emerged to push perceptual quality. HI-Diff~\cite{chen2023hidiff} integrates hierarchical latent priors into a diffusion framework for deblurring, and BlurDM~\cite{he2025blurdm} embeds the blur formation process into a dual-diffusion scheme (jointly diffusing noise and blur) and serves as a prior for downstream deblurring networks.
A parallel line adapts large pretrained generative models for image restoration. DiffBIR~\cite{lin2024diffbir} restores blind degradations using the Stable Diffusion~\cite{rombach2022high} prior, and Diff-Plugin~\cite{liu2024diffplugin} provides plug-ins for diffusion-based low-level tasks.
Most recently, several methods adapt the same prior specifically for deblurring. 
UID-Diff~\cite{cheng2025uiddiff} addresses unpaired deblurring with a decoupled diffusion model on top of a refined VAE~\cite{kingma2013auto}, and the concurrent FideDiff~\cite{liu2025fidediff} adapts Stable Diffusion through consistency training for one-step inference.
However, all these methods predict only a single sharp frame at the temporal center of the exposure, leaving users unable to pick any other moment encoded in the blur.

\paragraph{Blur-to-video.}
Beyond single-frame deblurring, blur-to-video recovers an entire sharp sequence from a single blurry image, inverting the temporal integration of the exposure.
\cite{jin2018learning} first formulated this task with an ordering-invariant loss that accepts both forward and reversed sequences as valid targets, declaring the directional ambiguity unresolvable.
Subsequent methods reduce the ambiguity via handcrafted motion priors~\cite{purohit2019bringing,argaw2021restoration} with limited out-of-distribution robustness.
MotionETR~\cite{zhang2021motionetr} estimates per-pixel motion offsets and warps a predicted sharp image to produce the sequence.
Animation-from-Blur~\cite{zhong2022animation} models motion ambiguity as multi-modal and conditions decomposition on a quantized motion-direction guidance (four discrete optical-flow directions), generating multiple plausible decompositions.
HyperCUT~\cite{pham2023hypercut} learns a hyperplane in latent space that separates forward from reversed sequences, using the vector side to assign temporal order.
Most recently, Blur2Vid~\cite{tedla2025blur2vid} fine-tunes a video diffusion transformer and treats directional ambiguity by sampling, producing plausible draws rather than a single committed reconstruction.
Despite these advances, all methods predict sharp content directly, rather than modeling the integral structure that defines motion blur.

\paragraph{Average-velocity Flow Matching.}
Flow Matching~\cite{lipman2023flow,liu2023rectified,albergo2023stochastic} is a generative modeling framework that learns to transform a simple prior distribution into a complex data distribution.
This is achieved by training a network to predict the instantaneous velocity field along a continuous transport path between the two distributions.
MeanFlow~\cite{geng2025meanflow} instead trains a network on the \emph{average velocity} over a time interval, which encodes the trajectory of the entire interval in a single prediction and lets one model handle arbitrary-interval queries.
Subsequently, SplitMeanFlow~\cite{guo2025splitmeanflow} introduces an algebraic additivity identity as a training objective, and pixel MeanFlow (pMF)~\cite{lu2026one} predicts an image-like quantity for one-step generation without a latent space.
Unlike these generative models, we adopt only the interval-averaging structure, predicting the temporal average of sharp frames over a sub-interval of the exposure in the latent space of a frozen VAE.

\vspace{-1em}
\section{Method}
\vspace{-0.5em}
\subsection{Preliminary: MeanFlow}
\vspace{-0.5em}
MeanFlow~\cite{geng2025meanflow} is a generative modeling framework that introduces the notion of \emph{average velocity} to characterize flow fields, in contrast to the instantaneous velocity modeled by Flow Matching.
Given a flow trajectory with state $z_t$ at time $t$, the average velocity over $[s, t]$ is the time-average of the instantaneous velocity $v$:
\begin{equation}
    u(z_t, s, t) \triangleq \frac{1}{t-s}\int_s^t v(z_\tau, \tau)\,\mathrm{d}\tau.
    \label{eq:meanflow_integral}
\end{equation}
Multiplying Eq.~\ref{eq:meanflow_integral} by $(t-s)$ and differentiating with respect to $t$ (with fixed $s$), we obtain the following relation via the fundamental theorem of calculus, referred to as the \emph{MeanFlow Identity}:
\begin{equation}
    u(z_t, s, t) = v(z_t, t) - (t-s)\tfrac{d}{dt}u(z_t, s, t).
    \label{eq:meanflow_identity}
\end{equation}

Since the true average $u$ is unobserved, MeanFlow trains a network $u_\theta$ by regressing it to a target derived from the MeanFlow Identity (a \emph{consistent estimator} in the sense that it equals $u$ whenever $u_\theta$ does):
\begin{equation}
    \mathcal{L}_\text{MeanFlow} = \mathbb{E}\Big\| u_\theta(z_t, s, t) - \mathrm{sg}\big(\hat{u}_\text{id}\big) \Big\|_2^2,
    \quad
    \hat{u}_\text{id} = v(z_t, t) - (t-s)\tfrac{d}{dt}u_\theta(z_t, s, t),
    \label{eq:meanflow_loss}
\end{equation}
where $\mathrm{sg}(\cdot)$ denotes stop-gradient and the instantaneous prediction emerges as the zero-interval limit: $\lim_{s \to t} u(z_t, s, t) = v(z_t, t)$.

While effective, computing the target $\hat{u}_\text{id}$ in Eq.~\ref{eq:meanflow_loss} requires a Jacobian-vector product (JVP) through the network to evaluate $\tfrac{d}{dt}u_\theta$, which increases per-step compute and memory.
SplitMeanFlow~\cite{guo2025splitmeanflow} addresses this by deriving an \emph{algebraic} counterpart of the MeanFlow Identity: splitting the integration domain at any $m \in [s, t]$ in Eq.~\ref{eq:meanflow_integral} yields,
\begin{equation}
    u(z_t, s, t) = \omega\, u(z_m, s, m) + (1{-}\omega)\, u(z_t, m, t), \quad \omega = \tfrac{m-s}{t-s},
    \label{eq:meanflow_additivity}
\end{equation}
which follows directly from $\int_s^t = \int_s^m + \int_m^t$ and recovers the differential MeanFlow Identity in the limit $m\!\to\!t$. 
SplitMeanFlow uses Eq.~\ref{eq:meanflow_additivity} as a JVP-free training target, reducing per-step compute and memory.

\vspace{-0.5em}
\subsection{Continuous-Time Blur Reformulation}
\label{subsec:analogy}
\vspace{-0.5em}
Conventional deblurring methods predict only a single sharp frame at the temporal center of the exposure, collapsing the exposure to one moment.
In contrast, to enable queries over arbitrary sub-intervals of the exposure, we first generalize the blur formation in Eq.~\ref{eq:blur_concept} from the full exposure interval $[0, T]$ to an arbitrary sub-interval $[s, t]$.
Formally, for any $0 \leq s < t \leq T$, the interval-mean blur is expressed as:
\begin{equation}
    B(s, t) \triangleq \frac{1}{t-s} \int_s^t S(\tau)\,\mathrm{d}\tau,
    \label{eq:blur_subinterval}
\end{equation}
where $S(\tau)$ is the sharp frame at time $\tau$ and $B(s, t)$ is the average sharp content over $[s, t]$. 
Throughout, we assume uniform temporal integration over a known exposure window, as in standard blur synthesis~\cite{nah2017deep}; Appendix~\ref{app:limitations} discusses other regimes.
This approach unifies the time-continuous representation of average sharp content $B(s,t)$, thereby generalizing conventional methods that recover a mid-exposure sharp image from a blurred observation $B(0,T)$, by interpreting the sharp frame $S(T/2)$ as the zero-interval limit $\lim_{t\to T/2} B(T/2, t)$.

\paragraph{Analogy with MeanFlow.}
We focus on the formal analogy between the interval-mean formulation of blur in Eq.~\ref{eq:blur_subinterval} and the average velocity in MeanFlow (Eq.~\ref{eq:meanflow_integral}): both are time-averages of an underlying quantity over a temporal interval, with the correspondence $B \leftrightarrow u$ and $S \leftrightarrow v$.
The analogy holds at the level of the averaging operator rather than the integrand: MeanFlow averages a velocity, whereas $B(s,t)$ averages sharp images, so the identities below carry over while MeanFlow's stochastic noise-to-data sampling does not.
Applying the same derivation as the \emph{MeanFlow Identity}, we obtain the analogous \emph{blur differential identity}:
\begin{equation}
    B(s, t) = S(t) - (t-s)\tfrac{d}{dt}B(s, t).
    \label{eq:blur_identity}
\end{equation}
Likewise, the splitting argument behind Eq.~\ref{eq:meanflow_additivity} carries over to the blur integral, yielding the \emph{blur additivity identity}: for any $m \in [s, t]$,
\begin{equation}
    B(s, t) = \omega\, B(s, m) + (1-\omega)\, B(m, t), \quad \omega = \tfrac{m-s}{t-s},
    \label{eq:blur_additivity}
\end{equation}
which admits a clean physical reading: a blur over an interval is the convex combination of two adjacent sub-exposure blurs.
The blur differential identity also implies a useful boundary case: as the interval shrinks to zero, the blur converges to the sharp frame at that instant:
\begin{equation}
    \lim_{t \to \tau} B(\tau, t) = S(\tau),
    \label{eq:blur_lim_sharp}
\end{equation}
making single-image deblurring at an arbitrary continuous time $\tau$ the zero-interval limit of the same formulation, and enabling continuous-time \textbf{``pick-a-moment''} recovery from a single blurred image, as well as continuous-time sharp video generation.

\paragraph{Architecture and conditioning.}
We instantiate $f_\theta$ as a diffusion transformer built on the FLUX.2-klein backbone~\cite{flux2024}, which consists of a frozen VAE with encoder $\mathcal{E}$ and decoder $\mathcal{D}$, along with a transformer $f_\theta$. 
The input blurry image $x_{\text{blur}}$ is encoded into the latent space as $z_{\text{blur}} = \mathcal{E}(x_{\text{blur}})$, and  $f_\theta$ is parameterized as a residual prediction: the transformer takes $(z_\text{blur}, s, t)$ as input and outputs a latent displacement that is added to  $z_\text{blur}$ to produce $f_\theta(z_\text{blur}, s, t)$.
To inject interval information, we replace FLUX's original time conditioning with an interval conditioner: the triple $(s, t, t{-}s)$ is sinusoidally embedded~\cite{vaswani2017attention} and projected by a shared MLP, then injected through the same conditioning pathway as the original timestep.

\begin{figure*}[t]
    \centering
    \includegraphics[width=\textwidth]{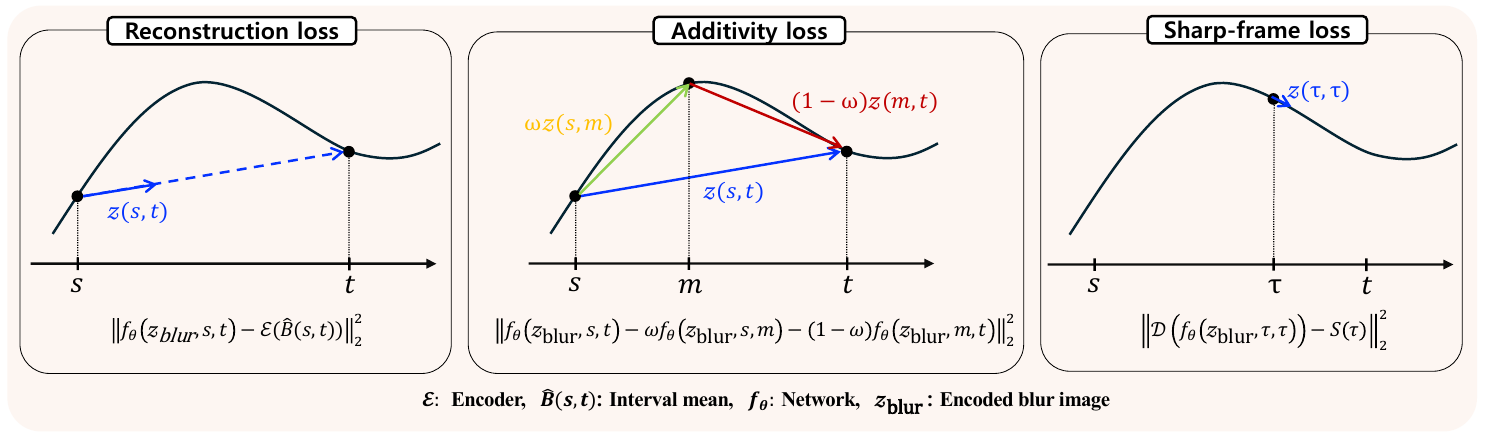}
    \vspace{-1.5em}
    \caption{Conceptual illustration of the training strategy of PickMoment.}
    \vspace{-1.5em}
    \label{fig:mainfigure}
\end{figure*}

\vspace{-0.5em}
\subsection{Training Strategy}
\label{sec:method:training}
\vspace{-0.5em}
We train $f_\theta$ by regressing it toward target estimators of the true interval mean $B(s, t)$, following the training principle of MeanFlow in Eq.~\ref{eq:meanflow_loss}. 
As illustrated in Fig.~\ref{fig:mainfigure}, we employ three complementary supervision signals, each derived from the integral formulation in Eq.~\ref{eq:blur_subinterval}: an empirical mean computed from available subframes, an additivity-based self-consistency across sub-intervals, and a zero-interval boundary condition motivated by the MeanFlow analogy that anchors sharp-frame predictions.

\paragraph{Reconstruction loss (empirical supervision from subframes).}
Conventional motion-blur datasets directly provide the sharp subframes $\{S(\tau_i)\}$ that correspond to each blurry observation.
This gives a direct ground-truth signal not available in MeanFlow's setting: we compute the empirical interval mean by averaging the subframes that fall within $[s, t],$
\begin{equation}
    \hat{B}(s, t) = \frac{1}{M_{s,t}} \sum_{\tau_i = s}^{t} S(\tau_i),
    \quad 
    \label{eq:Bhat_num}
\end{equation}
where $\hat{B}$ denotes the interval mean, and $M_{s,t}$ is the number of subframes within the interval $[s, t]$. 
Since $\hat{B}$ provides an empirical estimate of $B(s,t)$ from available subframes, we follow the MeanFlow analogy as discussed in Sec.~\ref{subsec:analogy}.
Accordingly, regressing $f_\theta$ to its encoded counterpart mirrors the form of $\mathcal{L}_\text{MeanFlow}$. 
\begin{equation}
    \mathcal{L}_\text{recon} = \mathbb{E}\Big\| f_\theta(z_\text{blur}, s, t) - \mathrm{sg}\!\big(\mathcal{E}(\hat{B}(s, t))\big) \Big\|_2^2,
    \label{eq:loss_recon}
\end{equation}
with $(s, t)$ sampled uniformly from the set of subframe-aligned pairs ${(\tau_i, \tau_j) : 0 \leq i \leq j \leq M{-}1}$ (Appendix~\ref{app:training}).

\paragraph{Additivity loss (self-consistency across sub-intervals).}
While $\mathcal{L}_\text{recon}$ supervises sampled interval queries, it provides no structural guarantee that $f_\theta$ behaves as a true interval mean function. 
We therefore enforce the blur additivity identity (Eq.~\ref{eq:blur_additivity}) as a self-consistency constraint.
Substituting $f_\theta$ into the right-hand side of Eq.~\ref{eq:blur_additivity} yields the additivity-form training target (the blur-domain analog of SplitMeanFlow's self-consistency target~\cite{guo2025splitmeanflow}), evaluable with two forward passes through $f_\theta$ without any time-derivative computation:
\begin{equation}
    \mathcal{E}(\hat{B}(s,t)) = \omega\, f_\theta(z_\text{blur}, s, m) + (1-\omega)\, f_\theta(z_\text{blur}, m, t).
\end{equation}
Regressing $f_\theta(z_\text{blur}, s, t)$ to this target yields the additivity loss:
\begin{equation}
    \mathcal{L}_\text{add} = \mathbb{E}\Big\| f_\theta(z_\text{blur}, s, t) - \omega f_\theta(z_\text{blur}, s, m) - (1-\omega) f_\theta(z_\text{blur}, m, t) \Big\|_2^2
    \label{eq:loss_add},
\end{equation}
which enforces consistency across overlapping intervals by sampling $m$ uniformly from the subframe timestamps strictly between $s$ and $t$, $\{\tau_k : s < \tau_k < t\}$, thereby propagating the point-wise supervision of $\mathcal{L}_\text{recon}$ into a coherent function over the full $[s, t]$ domain.\footnote{Strictly, the additivity in Eq.~\ref{eq:blur_additivity} holds in pixel space; applying it to the latent outputs of $f_\theta$ (Eq.~\ref{eq:loss_add}) relies on the approximate linearity of the VAE encoder over perceptually similar inputs, and acts as an effective structural prior in practice (validated in Appendix~\ref{app:additivity_validity}). A pixel-space variant would require additional decoder evaluations at each training step, incurring substantial computational overhead and increased training time.}

\paragraph{Sharp-frame loss (the deblurring boundary case).}
Single-image deblurring at an arbitrary timestamp $\tau$ corresponds to the zero-interval query $s = t = \tau$ of $f_\theta$, with target $S(\tau)$ as defined in Eq.~\ref{eq:blur_lim_sharp}. 
We supervise this case directly using the GT subframe at $\tau$, with a pixel-space $\ell_1$ distance and a perceptual EA-LPIPS distance~\cite{liu2025fidediff} computed after decoding, to encourage sharp, photo-realistic outputs:
\begin{equation}
    \mathcal{L}_\text{sharp} = \mathbb{E}_\tau\Big[\big\|\mathcal{D}(f_\theta(z_\text{blur}, \tau, \tau)) - S(\tau)\big\|_1 + \lambda_\text{perc}\,\mathrm{EA\text{-}LPIPS}\!\big(\mathcal{D}(f_\theta(z_\text{blur}, \tau, \tau)),\, S(\tau)\big)\Big],
    \label{eq:loss_sharp}
\end{equation}
with $\tau$ sampled uniformly from the subframe timestamps $\{\tau_i\}_{i=0}^{M-1}$.
This anchors the deblurring output directly with GT supervision, complementing the interval-level signals from $\mathcal{L}_\text{recon}$ and $\mathcal{L}_\text{add}$.

\paragraph{Total objective.} 
The complete loss combines the three terms: 
\begin{equation}
    \mathcal{L}_\text{total} = \mathcal{L}_\text{recon} + \lambda_\text{add} \mathcal{L}_\text{add} + \lambda_\text{sharp} \mathcal{L}_\text{sharp}.
\end{equation} 
The auxiliary weights are set to $(\lambda_\text{add}, \lambda_\text{sharp}, \lambda_\text{perc}) = (0.1, 1.0, 2.0)$, prioritizing the empirical reconstruction signal $\mathcal{L}_\text{recon}$; the full warm-up schedule is provided in Appendix~\ref{app:training}.

\vspace{-0.5em}
\subsection{Inference}
\vspace{-5px}
\label{sec:method:inference}
A single trained $f_\theta$ unifies three inference modes by varying the interval query, each evaluated in one forward pass without retraining or iterative sampling.
For \emph{single-image deblurring}, we query at the zero-interval limit $(s,t)=(T/2-\epsilon,\,T/2+\epsilon)$ to recover the center sharp frame, where $\epsilon$ denotes a small constant.
For \emph{blur-to-video} with $N$ output frames, we query at $N$ uniformly spaced zero-intervals $\{(\tau_i-\epsilon,\,\tau_i+\epsilon)\}_{i=0}^{N-1}$ with $\tau_i = i/(N-1)$, recovering the sharp sequence over the exposure.
For \emph{pick-a-moment} recovery, we query at $(\tau-\epsilon,\,\tau+\epsilon)$ for any user-specified $\tau\in[0,T]$.
All three modes share the same parameters and produce deterministic outputs given the same blurry input.
Here, deterministic refers to the inference procedure, which samples no noise as in~\cite{wu2024osediff,liu2025fidediff}; although built on a pretrained generative prior, PickMoment does not retain the sampling-based generative capability of stochastic inference.
We use $\epsilon = 0.01$ as the default zero-interval width.

\vspace{-0.5em}
\section{Experiments}
\label{sec:exp}

\vspace{-0.5em}
\subsection{Experimental Setup}
\vspace{-0.5em}
\label{sec:exp:setup}
\paragraph{Datasets.} 
We train on the GoPro dataset~\cite{nah2017deep}.
Each blurry image is synthesized by averaging 7--13 consecutive sharp frames with a known temporal order over the exposure window, and we use all of these subframes as supervision targets during training.
We train only on the GoPro training split and apply the same model to all evaluation datasets without any per-dataset fine-tuning.
We conduct experiments on GoPro~\cite{nah2017deep}, HIDE~\cite{shen2019hide}, RealBlur-J, and RealBlur-R~\cite{rim2020realblur} for single-frame evaluation. 
For blur-to-video evaluation, we use GoPro-7, the 7-frame variant of the GoPro test split established by Jin et al.~\cite{jin2018learning} and adopted by subsequent work~\cite{tedla2025blur2vid}, where each blurred image is paired with seven sharp frames spanning the exposure.
\vspace{-5px}
\paragraph{Metrics.} 
For single-image deblurring, we report PSNR, SSIM~\cite{wang2004ssim}, LPIPS~\cite{zhang2018lpips}, and DISTS~\cite{ding2020dists} computed against the center sharp frame of each blur exposure, following the FideDiff~\cite{liu2025fidediff} setup.
For blur-to-video, we additionally report the bidirectional patch-based variants PSNR$_p$, SSIM$_p$, and LPIPS$_p$ according to Blur2Vid~\cite{tedla2025blur2vid}, which take a per-patch minimum between the prediction and its temporally-reversed counterpart against the ground-truth sequence to account for the directional ambiguity of motion-blur inversion, together with FVD~\cite{unterthiner2019fvd} and the end-point error (EPE) of optical flow computed via RAFT~\cite{teed2020raft}. 
All metrics are measured at $640\!\times\!360$ resolution to ensure a fair comparison with prior blur-to-video methods~\cite{jin2018learning,zhang2021motionetr,tedla2025blur2vid}.

\paragraph{Training details.}
We optimize with Adam~\cite{kingma2015adam} for a total of 100K iterations.
The learning rate is linearly warmed up to $10^{-4}$ over the first 2K steps and then cosine-annealed~\cite{loshchilov2017sgdr} to $10^{-5}$.
We use bf16 mixed precision with gradient accumulation of 2 steps.
Following Restormer~\cite{zamir2022restormer}, we adopt progressive learning where the (patch size, batch size) pairs are updated to $[(256^2, 8), (384^2, 4), (512^2, 2)]$ at iterations $[30\text{K}, 60\text{K}, 100\text{K}]$.
Training is performed on a single NVIDIA RTX A6000 GPU and takes around 4 days.
Full hyperparameters are given in Appendix~\ref{app:training}.

\begin{table}[t]
\centering
\caption{
Quantitative comparison for single-image deblurring. 
\textbf{Bold} indicates the best result and \underline{underline} indicates the second-best result among generative-prior-based methods.
}
\label{tab:main_deblur}
\resizebox{0.99\textwidth}{!}{%
\begin{tabular}{l|l| cccc||ccccc}
\toprule
\rowcolor{HeaderBlue}
 & & \multicolumn{4}{c||}{Restoration-based} & \multicolumn{5}{c}{Generative-prior-based} \\
\rowcolor{HeaderBlue}
Dataset & Metric
  & Restormer~\cite{zamir2022restormer} & HI-Diff~\cite{chen2023hidiff} & UFPNet~\cite{fang2023ufpnet} & AdaRevD~\cite{mao2024adarevd} & DiffBIR~\cite{lin2024diffbir} & OSEDiff~\cite{wu2024osediff} & Diff-Plugin~\cite{liu2024diffplugin} & FideDiff~\cite{liu2025fidediff}
  & Ours \\
\midrule
\multirow{4}{*}{GoPro}
  & PSNR$\uparrow$  & 32.92 & 33.33 & 34.09 & 34.60 & 26.15 & 24.34 & 22.88 & \underline{28.79} & \textbf{29.43} \\
  & SSIM$\uparrow$  & 0.9611 & 0.9642 & 0.9686 & 0.9716 & 0.8377 & 0.8228 & 0.7798 & \underline{0.9148} & \textbf{0.9253} \\
  & LPIPS$\downarrow$ & 0.0841 & 0.0799 & 0.0764 & 0.0712 & 0.2366 & 0.1738 & 0.2332 & \underline{0.0831} & \textbf{0.0802} \\
  & DISTS$\downarrow$ & 0.0724 & 0.0710 & 0.0666 & 0.0672 & 0.1460 & 0.0834 & 0.1166 & \underline{0.0525} & \textbf{0.0429} \\
\midrule
\multirow{4}{*}{HIDE}
  & PSNR$\uparrow$  & 31.22 & 31.46 & 31.74 & 32.35 & 25.94 & 23.20 & 21.94 & \underline{27.28} & \textbf{27.52} \\
  & SSIM$\uparrow$  & 0.9423 & 0.9446 & 0.9471 & 0.9525 & 0.8216 & 0.7611 & 0.7272 & \underline{0.8775} & \textbf{0.8891} \\
  & LPIPS$\downarrow$ & 0.1082 & 0.1055 & 0.0931 & 0.0872 & 0.2091 & 0.2208 & 0.2732 & \underline{0.1068} & \textbf{0.1046} \\
  & DISTS$\downarrow$ & 0.0730 & 0.0725 & 0.0676 & 0.0666 & 0.1246 & 0.1001 & 0.1411 & \underline{0.0647} & \textbf{0.0553} \\
\midrule
\multirow{4}{*}{RealBlur-J}
  & PSNR$\uparrow$  & 28.96 & 29.15 & 29.87 & 30.12 & 26.92 & 26.83 & 25.77 & \textbf{28.96} & \underline{28.94} \\
  & SSIM$\uparrow$  & 0.8786 & 0.8898 & 0.8840 & 0.8945 & 0.7450 & 0.8004 & 0.7711 & \underline{0.8695} & \textbf{0.8757} \\
  & LPIPS$\downarrow$ & 0.1561 & 0.1470 & 0.1441 & 0.1408 & 0.2587 & 0.1793 & 0.2055 & \textbf{0.1142} & \underline{0.1184} \\
  & DISTS$\downarrow$ & 0.1112 & 0.1050 & 0.1101 & 0.1037 & 0.1599 & 0.1198 & 0.1355 & \textbf{0.0800} & \underline{0.0841} \\
\midrule
\multirow{4}{*}{RealBlur-R}
  & PSNR$\uparrow$  & 36.19 & 36.28 & 36.25 & 36.53 & 32.60 & 33.54 & 32.64 & \underline{36.01} & \textbf{36.08} \\
  & SSIM$\uparrow$  & 0.9572 & 0.9583 & 0.9528 & 0.9570 & 0.8493 & 0.9056 & 0.8496 & \underline{0.9424} & \textbf{0.9510} \\
  & LPIPS$\downarrow$ & 0.0608 & 0.0602 & 0.0615 & 0.0621 & 0.3388 & 0.1057 & 0.1422 & \textbf{0.0584} & \underline{0.0612} \\
  & DISTS$\downarrow$ & 0.0833 & 0.0814 & 0.0854 & 0.0846 & 0.2576 & 0.1322 & 0.1673 & \underline{0.0862} & \textbf{0.0833} \\
\bottomrule
\end{tabular}%
}
\vspace{-1em}
\end{table}

\vspace{-0.7em}
\subsection{Single-Image Deblurring Results}
\label{sec:exp:deblur}
\vspace{-5px}
We compare PickMoment against representative single-image deblurring methods spanning two families: \emph{restoration-based} approaches (Restormer~\cite{zamir2022restormer}, HI-Diff~\cite{chen2023hidiff}, UFPNet~\cite{fang2023ufpnet}, AdaRevD~\cite{mao2024adarevd}) and \emph{generative-prior-based} ones (DiffBIR~\cite{lin2024diffbir}, OSEDiff~\cite{wu2024osediff}, Diff-Plugin~\cite{liu2024diffplugin}, FideDiff~\cite{liu2025fidediff}).
For deblurring, PickMoment is evaluated using the zero-interval limit $(s,t)=(T/2-\epsilon,\,T/2+\epsilon)$.
\vspace{-5px}
\paragraph{Quantitative results.}
Within the family of generative-prior-based methods, Table~\ref{tab:main_deblur} shows that PickMoment achieves state-of-the-art performance on GoPro and HIDE across all four metrics, surpassing FideDiff by $+0.64$ dB PSNR on GoPro and $+0.24$ dB on HIDE.
Results on RealBlur-J/R are competitive with the closely matched FideDiff, while PickMoment achieves better performance on 12 out of 16 evaluation metrics overall. 
Transformer-based baselines such as AdaRevD~\cite{mao2024adarevd} achieve higher per-pixel PSNR but predict only a single sharp frame and do not natively support blur-to-video.

\begin{figure*}[t]
    \centering
    \includegraphics[width=\textwidth,]{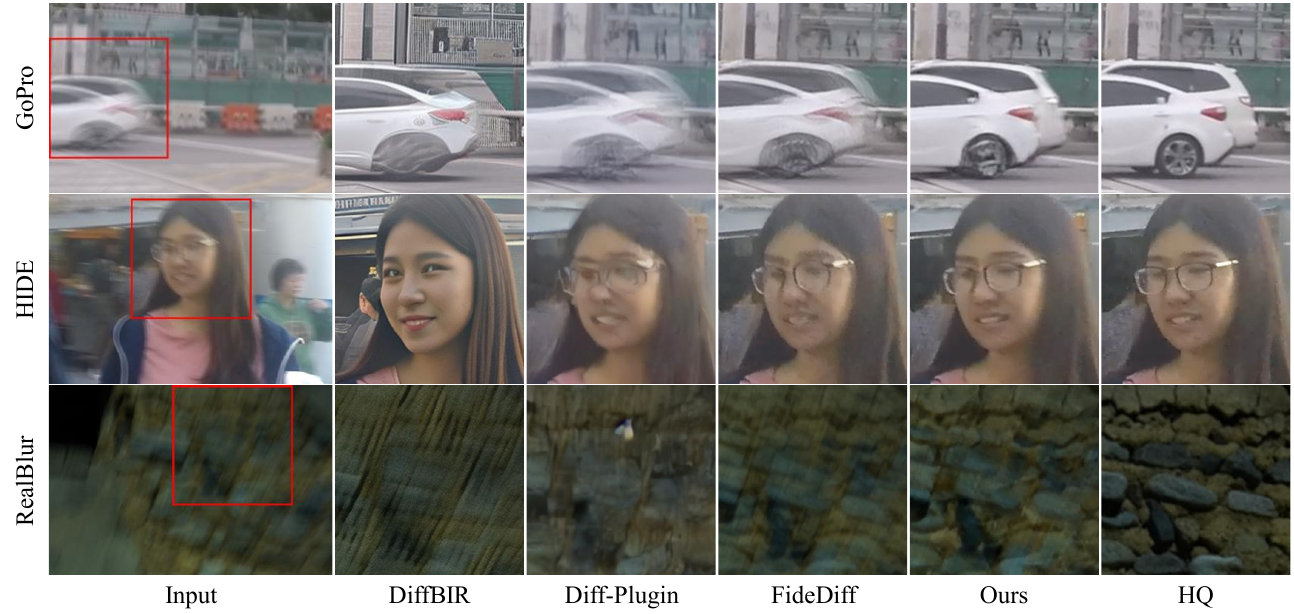}
    \vspace{-1.0em}
    \caption{Qualitative comparison for single-image deblurring on GoPro, HIDE, and RealBlur-J (top to bottom). Best viewed zoomed in.}
    \vspace{-1.0em}
    \label{fig:qual_deblur}
\end{figure*}

\vspace{-0.5em}
\paragraph{Qualitative results.}
Fig.~\ref{fig:qual_deblur} compares deblurred outputs on representative samples from GoPro, HIDE, and RealBlur.
PickMoment restores sharper local structures than existing diffusion-based baselines, especially around vehicle boundaries, facial details, and fine texture patterns. The results show clearer edges, reduced residual blur, and fewer structural distortions. Additional visual comparisons are provided in Appendix~\ref{app:visuals}.

\vspace{-0.7em}
\subsection{Blur-to-Video Results}
\label{sec:exp:sequence}
\vspace{-0.5em}
We compare PickMoment against prior blur-to-video methods that recover an in-exposure sharp sequence from a single blurred input: Jin et al.~\cite{jin2018learning}, MotionETR~\cite{zhang2021motionetr}, and Blur2Vid~\cite{tedla2025blur2vid}, which is the state-of-the-art video-diffusion-based approach. 
Given a blurred input, PickMoment reconstructs the full exposure-time sequence through predictions of $f_\theta$ at $N$ uniformly spaced intervals (Sec.~\ref{sec:method:inference}).

\vspace{-0.7em}
\paragraph{Quantitative results.}
Table~\ref{tab:main_seq} shows that PickMoment achieves the highest per-frame fidelity, surpassing Blur2Vid by $+1.23$ dB in PSNR$_p$ and reducing LPIPS$_p$ by a relative margin of $28\%$. 
Blur2Vid retains an advantage on temporal smoothness (FVD, EPE), reflecting an architectural trade-off: deterministic per-interval prediction maximizes per-frame quality, whereas video-diffusion sampling provides implicit temporal regularization. 
PickMoment needs one forward pass per output frame, issued as a single batch ($7.90$\,s for a 7-frame sequence), in contrast to Blur2Vid's iterative sampling, yielding roughly a $25\times$ inference speedup (Appendix~\ref{app:inference_cost}).
A warping-based check (Appendix~\ref{app:temporal}) shows that our frame-to-frame change is close to that of the ground truth (warp L1 $0.0089$ vs.\ $0.0096$), whereas Blur2Vid is smoother than the ground truth ($0.0072$).
Across all 1{,}744 GoPro-7 clips, PickMoment also commits to one motion direction rather than blending both (median forward/reverse PSNR gap $9.2$ dB), which we report as an empirical characterization rather than a guarantee.

\begin{table*}[t]
\centering
\begin{minipage}[t]{0.47\linewidth}
\centering
\caption{Quantitative comparison for blur-to-video on GoPro-7.}
\label{tab:main_seq}
\resizebox{\linewidth}{!}{%
\begin{tabular}{lccccc}
\toprule
\rowcolor{HeaderBlue}
Method & PSNR$_p$$\uparrow$ & SSIM$_p$$\uparrow$ & LPIPS$_p$$\downarrow$ & FVD$\downarrow$ & EPE$\downarrow$ \\
\midrule
Jin et al.~\cite{jin2018learning}   & 25.23 & 0.8190 & 0.084  & 235.53              & 3.38              \\
MotionETR~\cite{zhang2021motionetr} & 26.54 & 0.8825 & 0.015  & 94.90               & 1.46              \\
Blur2Vid~\cite{tedla2025blur2vid}   & \underline{30.01} & \underline{0.9359} & \underline{0.012}  & \textbf{21.46}      & \textbf{0.39}     \\
\midrule
Ours & \textbf{31.24} & \textbf{0.9388} & \textbf{0.0086} & \underline{27.40} & \underline{0.86} \\
\bottomrule
\end{tabular}}
\end{minipage}
\hfill
\begin{minipage}[t]{0.52\linewidth}
\centering
\caption{Generalization ablation on unseen timestamps and $\mathbf{B \rightarrow B}$ denotes our proposed method.
}
\label{tab:abl_time}
\resizebox{\linewidth}{!}{%
\begin{tabular}{ll cccccc}
\toprule
\rowcolor{HeaderBlue}
Method & Metric & $\tau=2$ & $\tau=4$ & $\tau=6$ & $\tau=8$ & $\tau=10$ & Mean \\
\midrule
\multirow{2}{*}{$\mathbf{B \rightarrow S}$}
  & PSNR$\uparrow$ & 26.51 & 27.47 & 28.42 & 27.41 & 26.52 & 27.27 \\
  & SSIM$\uparrow$ & 0.8273 & 0.8450 & 0.8626 & 0.8450 & 0.8288 & 0.8418 \\
\midrule
\multirow{2}{*}{$\mathbf{B \rightarrow B}$}
  & PSNR$\uparrow$ & \textbf{27.34} & \textbf{28.34} & \textbf{29.44} & \textbf{28.30} & \textbf{27.55} & \textbf{28.19} \\
  & SSIM$\uparrow$ & \textbf{0.8447} & \textbf{0.8682} & \textbf{0.8860} & \textbf{0.8681} & \textbf{0.8560} & \textbf{0.8646} \\
\bottomrule
\end{tabular}}
\end{minipage}
\end{table*}

\begin{figure*}[t]
    \centering
    \vspace{-1.0em}
    \includegraphics[width=\textwidth]{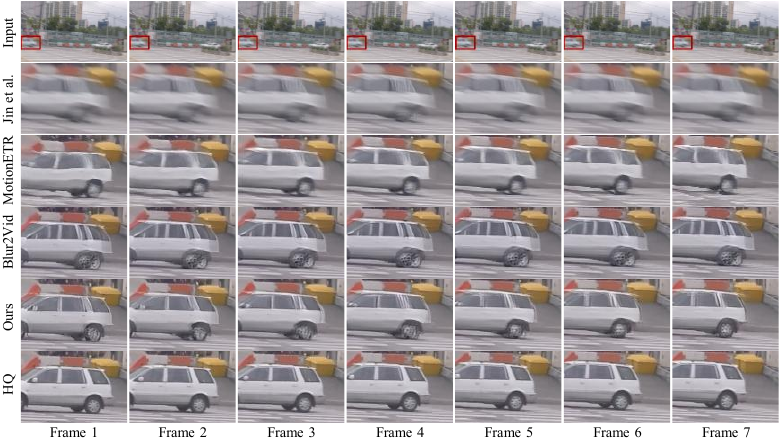}
    \vspace{-1.5em}
    \caption{Qualitative comparison for blur-to-video on GoPro-7. Best viewed zoomed in.}
    \label{fig:qual_seq}
    \vspace{-0.5em}
\end{figure*}

\vspace{-0.7em}
\paragraph{Qualitative results.}
Fig.~\ref{fig:qual_seq} compares recovered in-exposure sequences against Blur2Vid~\cite{tedla2025blur2vid} on GoPro-7.
Jin et al.~\cite{jin2018learning} leaves substantial blur and recovers little sharp structure, while MotionETR noticeably distorts the vehicle geometry. 
Blur2Vid produces more plausible frames, but residual blur and softened details remain around the vehicle body and wheels. 
PickMoment better restores sharp object boundaries and local textures, yielding visually clearer and more structurally faithful in-exposure frames. Additional visual comparisons are provided in Appendix~\ref{app:visuals}.

\vspace{-1.0em}
\subsection{Ablation Studies}
\label{sec:exp:ablation}
\vspace{-0.5em}

We conduct three ablations to study (i) generalization to unseen timestamps, (ii) the choice of supervision target, and (iii) the contribution of each loss term.
All variants share the same backbone, training data, and training budget, and are trained using only $\mathcal{L}_\text{recon}$ in the latent space unless stated otherwise. 
This isolates the effect of the supervision target, while the impact of the loss design is analyzed separately in the loss ablation study. 
Three baselines are used: 
$(\mathbf{B \rightarrow S})$, which learns a blur-to-single-sharp-frame mapping by predicting a single sharp frame $S(\tau)$ at a given discrete timestamp $\tau$; 
$(\mathbf{B \rightarrow \{S(\tau_i)\}_{i=0}^{N-1}})$, which learns a blur-to-multi-sharp-frame mapping by predicting $N$ sharp frames simultaneously through a multi-channel output at predefined timestamps from a single blurred image, following~\cite{jin2018learning}; 
and \textbf{Ours ($\mathbf{B \rightarrow B}$)}, which predicts arbitrary interval-mean blurs from a blurred image. 
Because $\mathbf{B \rightarrow S}$ is a conventional sharp-frame predictor on the same pretrained backbone, Table~\ref{tab:abl_arch} also serves as a capacity-matched comparison, and a large pretrained prior is standard among the generative-prior-based baselines~\cite{lin2024diffbir,wu2024osediff,liu2024diffplugin,liu2025fidediff}.

\begin{table}[t]
\centering
\caption{Advantage of interval prediction on GoPro. $\mathbf{B \rightarrow B}$ denotes our proposed method.}
\label{tab:abl_arch}
\resizebox{\textwidth}{!}{%
\begin{tabular}{l cccc ccccc}
\toprule
\rowcolor{HeaderBlue}
 & \multicolumn{4}{c}{Deblurring} & \multicolumn{5}{c}{Blur-to-video} \\
\rowcolor{HeaderBlue}
Method & PSNR$\uparrow$ & SSIM$\uparrow$ & LPIPS$\downarrow$ & DISTS$\downarrow$ & PSNR$_p$$\uparrow$ & SSIM$_p$$\uparrow$ & LPIPS$_p$$\downarrow$ & FVD$\downarrow$ & EPE$\downarrow$ \\
\midrule
$\mathbf{B \rightarrow S}$ & 27.72 & 0.8944 & 0.1595 & 0.1149 & 29.16 & 0.8894 & 0.0292 & 59.51 & 1.15 \\
$\mathbf{B \rightarrow \{S(\tau_i)\}_{i=0}^{N-1}}$ & 27.76 & 0.8945 & 0.1604 & 0.1166 & 29.26 & 0.8933 & 0.0278 & \textbf{58.76} & 1.13 \\
\rowcolor{RowGray}
$\mathbf{B \rightarrow B}$  (ours)& \textbf{28.45} & \textbf{0.9080} & \textbf{0.1455} & \textbf{0.1043} & \textbf{29.78} & \textbf{0.9034} & \textbf{0.0271} & 59.29 & \textbf{1.02} \\
\bottomrule
\end{tabular}%
}

\centering
\caption{Loss-term ablation on GoPro. \checkmark{} indicates the loss is enabled.}
\label{tab:abl_loss}
\resizebox{\textwidth}{!}{%
\begin{tabular}{ccc cccc ccccc}
\toprule
\rowcolor{HeaderBlue}
\multicolumn{3}{c}{Losses} & \multicolumn{4}{c}{Deblurring} & \multicolumn{5}{c}{Blur-to-video} \\
\rowcolor{HeaderBlue}
$\mathcal{L}_\text{recon}$ & $\mathcal{L}_\text{add}$ & $\mathcal{L}_\text{sharp}$ & PSNR$\uparrow$ & SSIM$\uparrow$ & LPIPS$\downarrow$ & DISTS$\downarrow$ & PSNR$_p$$\uparrow$ & SSIM$_p$$\uparrow$ & LPIPS$_p$$\downarrow$ & FVD$\downarrow$ & EPE$\downarrow$ \\
\midrule
           &            & \checkmark & 29.06 & 0.9208 & 0.0812 & 0.0438 & 30.17 & 0.9186 & 0.0104 & 38.07 & 1.33 \\
\midrule
\checkmark &            &            & 28.45 & 0.9080 & 0.1455 & 0.1043 & 29.78 & 0.9034 & 0.0271 & 59.29 & 1.02 \\
\checkmark & \checkmark &            & 28.56 & 0.9096 & 0.1422 & 0.1020 & 29.95 & 0.9058 & 0.0260 & 58.91 & 0.92 \\
\checkmark &            & \checkmark & 29.24 & 0.9236 & \textbf{0.0787} & \textbf{0.0420} & 30.68 & 0.9255 & 0.0146 & 67.10 & 0.99 \\
\checkmark & \checkmark & \checkmark & \textbf{29.43} & \textbf{0.9253} & 0.0802 & 0.0429 & \textbf{31.24} & \textbf{0.9388} & \textbf{0.0086} & \textbf{27.40} & \textbf{0.86} \\
\bottomrule
\end{tabular}%
}
\end{table}

\vspace{-0.5em}
\paragraph{Generalization to unseen timestamps.}
In Table~\ref{tab:abl_time}, we evaluate whether interval supervision generalizes to timestamps unseen during training, as our model learns continuous-time blur representations rather than timestamp-specific predictions which are inaccessible to previous blur-to-video models. 
Both $\mathbf{B \rightarrow S}$ and Ours are trained on only the odd-indexed subframes%
, and we evaluate at the unseen even-indexed timestamps $\tau \in \{\frac{2}{11}, \frac{4}{11}, \frac{6}{11}, \frac{8}{11}, \frac{10}{11}\}$. 
Ours outperforms $\mathbf{B \rightarrow S}$ at every timestamp, with $+0.92$ dB Mean PSNR and $+0.023$ Mean SSIM values, indicating that interval supervision generalizes more smoothly across continuous time than direct sharp-frame prediction.
Beyond held-out GoPro timestamps, we evaluate zero-shot on X4K1000FPS~\cite{sim2021xvfi}, where averaging 33 consecutive 1000-fps frames leaves ground truth at 33 timestamps per exposure: PickMoment outperforms Blur2Vid at every timestamp, with a $1.35$ dB higher mean PSNR (Appendix~\ref{app:continuous_time}).
Querying the endpoint of a synthesized half-exposure blur, an input length unseen in training, also stays within 1 dB of center-frame deblurring on the full blur ($28.49$ vs.\ $29.43$ dB), which we view as supporting evidence for a continuous interval-mean field rather than proof. 

\vspace{-0.5em}
\paragraph{Advantage of continuous-time interval prediction.}
Table~\ref{tab:abl_arch} compares the three supervision targets on the standard evaluation.
Interval-mean supervision dominates on every metric except FVD, where $\mathbf{B \rightarrow \{S(\tau_i)\}}$ is marginally better, indicating that learning the integral yields stronger representations than learning direct sharp-frame prediction.

\begin{figure*}[t]
    \centering
    \includegraphics[width=\textwidth]{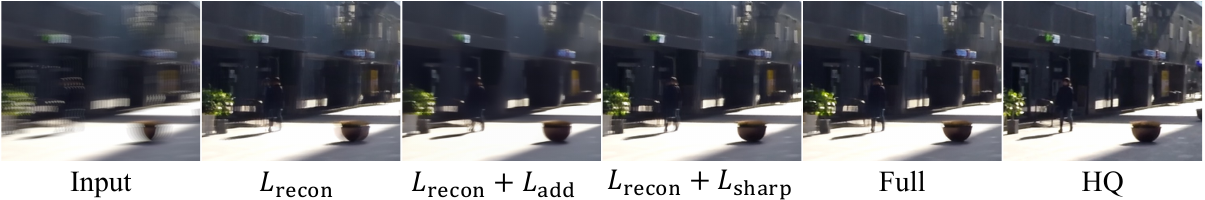}
    \vspace{-1.0em}
    \caption{Loss-term ablation visualization on GoPro-7 (frame 6 of 7).  Best viewed zoomed in.}
    \vspace{-0.5em}
    \label{fig:qual_loss}
\end{figure*}

\vspace{-10px}
\paragraph{Effectiveness of loss functions.}
We analyze the contribution of each loss term in the full objective (Table~\ref{tab:abl_loss}, Fig.~\ref{fig:qual_loss}).
Using $\mathcal{L}_\text{recon}$ alone produces coarse outputs (LPIPS $0.1455$), while $\mathcal{L}_\text{sharp}$ nearly halves the LPIPS value, accounting for most of the perceptual improvement.
$\mathcal{L}_\text{add}$ alone yields only marginal gains, but combined with $\mathcal{L}_\text{sharp}$ it reduces FVD from $67.10$ to $27.40$ and EPE from $0.99$ to $0.86$, showing that the two losses are complementary: $\mathcal{L}_\text{sharp}$ enforces per-frame sharpness while $\mathcal{L}_\text{add}$ propagates temporal coherence across the recovered sequence (Fig.~\ref{fig:qual_loss}).
In particular, $\mathcal{L}_\text{add}$ is especially effective for blur-to-video task, where temporal consistency is critical.
With $\mathcal{L}_\text{sharp}$ alone, deblurring stays close to the full objective while blur-to-video degrades (FVD $38.07$ vs.\ $27.40$), so the interval-mean formulation mainly improves temporal coherence.

\vspace{-1em}
\section{Conclusion}
\label{sec:conclusion}
\vspace{-0.5em}
We introduced PickMoment, a continuous-time reformulation of motion deblurring that predicts the interval-mean blur over arbitrary sub-intervals of the exposure.
A single deterministic model unifies single-image deblurring, blur-to-video, and pick-a-moment recovery as different queries, trained with three supervisions derived from the integral definition: an empirical reconstruction loss, an additivity loss for self-consistency across sub-intervals, and a sharp-frame loss at the zero-interval limit.
PickMoment achieves state-of-the-art single-image deblurring among generative-prior-based methods on GoPro and HIDE, competitive performance on RealBlur, and the highest per-frame fidelity on GoPro-7 blur-to-video, all with one forward pass per output frame.
The main remaining gap is temporal smoothness, where iterative video-diffusion sampling still provides stronger implicit regularization; combining our deterministic interval-mean prediction with iterative sampling is a natural next step.
We discuss further limitations in Appendix~\ref{app:limitations}.

\begin{ack}
This research was supported by Culture, Sports and Tourism R\&D Program through the Korea Creative Content Agency grant funded by the Ministry of Culture, Sports and Tourism in 2026 (Project Name: Development of AI-Based Animation Production Technology to Ensure Character and Scene Consistency and Continuity for Enhanced Efficiency and Quality, Project Number: RS-2026-25525207, Contribution Rate: 30\%).
This work was also supported by Institute of Information \& communications Technology Planning \& Evaluation (IITP) grant funded by the Korea government (MSIT) (No.2022-0-00156, Fundamental research on continual meta-learning for quality enhancement of casual videos and their 3D metaverse transformation), and IITP grant funded by the Korea government (MSIT) (No.RS-2020-II201373, Artificial Intelligence Graduate School Program (Hanyang University)).
\end{ack}

\small
\bibliographystyle{plainnat}
\bibliography{main}

\begin{thebibliography}{55}
\providecommand{\natexlab}[1]{#1}
\providecommand{\url}[1]{\texttt{#1}}
\expandafter\ifx\csname urlstyle\endcsname\relax
  \providecommand{\doi}[1]{doi: #1}\else
  \providecommand{\doi}{doi: \begingroup \urlstyle{rm}\Url}\fi

\bibitem[Albergo et~al.(2023)Albergo, Boffi, and
  Vanden-Eijnden]{albergo2023stochastic}
Michael~S. Albergo, Nicholas~M. Boffi, and Eric Vanden-Eijnden.
\newblock Stochastic interpolants: A unifying framework for flows and
  diffusions.
\newblock \emph{arXiv preprint arXiv:2303.08797}, 2023.

\bibitem[Argaw et~al.(2021)Argaw, Kim, Rameau, Zhang, and
  Kweon]{argaw2021restoration}
Dawit~Mureja Argaw, Junsik Kim, Francois Rameau, Chaoning Zhang, and In~So
  Kweon.
\newblock Restoration of video frames from a single blurred image with motion
  understanding.
\newblock In \emph{CVPR Workshops (NTIRE)}, 2021.

\bibitem[{Black Forest Labs}(2025)]{flux2024}
{Black Forest Labs}.
\newblock {FLUX.2}.
\newblock \url{https://github.com/black-forest-labs/flux2}, 2025.

\bibitem[Chen et~al.(2022)Chen, Chu, Zhang, and Sun]{chen2022simple}
Liangyu Chen, Xiaojie Chu, Xiangyu Zhang, and Jian Sun.
\newblock Simple baselines for image restoration.
\newblock In \emph{ECCV}, 2022.

\bibitem[Chen et~al.(2023)Chen, Zhang, Liu, Xia, Gu, Kong, and
  Yuan]{chen2023hidiff}
Zheng Chen, Yulun Zhang, Ding Liu, Bin Xia, Jinjin Gu, Linghe Kong, and Xin
  Yuan.
\newblock Hierarchical integration diffusion model for realistic image
  deblurring.
\newblock In \emph{NeurIPS}, 2023.

\bibitem[Cheng et~al.(2025)Cheng, Chen, Lu, and Yang]{cheng2025uiddiff}
Junhao Cheng, Wei-Ting Chen, Xi~Lu, and Ming-Hsuan Yang.
\newblock Unpaired deblurring via decoupled diffusion model.
\newblock \emph{arXiv preprint arXiv:2502.01522}, 2025.

\bibitem[Ding et~al.(2020)Ding, Ma, Wang, and Simoncelli]{ding2020dists}
Keyan Ding, Kede Ma, Shiqi Wang, and Eero~P. Simoncelli.
\newblock Image quality assessment: Unifying structure and texture similarity.
\newblock \emph{IEEE Transactions on Pattern Analysis and Machine
  Intelligence}, 2020.

\bibitem[Fang et~al.(2023)Fang, Wu, Dong, Li, Wu, and Shi]{fang2023ufpnet}
Zhenxuan Fang, Fangfang Wu, Weisheng Dong, Xin Li, Jinjian Wu, and Guangming
  Shi.
\newblock Self-supervised non-uniform kernel estimation with flow-based motion
  prior for blind image deblurring.
\newblock In \emph{CVPR}, 2023.

\bibitem[Fergus et~al.(2006)Fergus, Singh, Hertzmann, Roweis, and
  Freeman]{fergus2006removing}
Rob Fergus, Barun Singh, Aaron Hertzmann, Sam~T. Roweis, and William~T.
  Freeman.
\newblock Removing camera shake from a single photograph.
\newblock \emph{ACM Transactions on Graphics (SIGGRAPH)}, 2006.

\bibitem[Geng et~al.(2025)Geng, Deng, Bai, Kolter, and He]{geng2025meanflow}
Zhengyang Geng, Mingyang Deng, Xingjian Bai, J.~Zico Kolter, and Kaiming He.
\newblock Mean {Flows} for one-step generative modeling.
\newblock \emph{arXiv preprint arXiv:2505.13447}, 2025.

\bibitem[Guo et~al.(2025)Guo, Wang, Yuan, Cao, Chen, Chen, Huo, Zhang, Wang,
  Liu, and Wang]{guo2025splitmeanflow}
Yi~Guo, Wei Wang, Zhihang Yuan, Rong Cao, Kuan Chen, Zhengyang Chen, Yuanyuan
  Huo, Yang Zhang, Yuping Wang, Shouda Liu, and Yuxuan Wang.
\newblock {SplitMeanFlow}: Interval splitting consistency in few-step
  generative modeling.
\newblock \emph{arXiv preprint arXiv:2507.16884}, 2025.

\bibitem[He et~al.(2025)He, Tsai, Peng, Chen, Lin, and Lin]{he2025blurdm}
Jin-Ting He, Fu-Jen Tsai, Yan-Tsung Peng, Min-Hung Chen, Chia-Wen Lin, and
  Yen-Yu Lin.
\newblock {BlurDM}: A blur diffusion model for image deblurring.
\newblock In \emph{NeurIPS}, 2025.

\bibitem[Hu et~al.(2022)Hu, Shen, Wallis, Allen-Zhu, Li, Wang, Wang, and
  Chen]{hu2022lora}
Edward~J. Hu, Yelong Shen, Phillip Wallis, Zeyuan Allen-Zhu, Yuanzhi Li, Shean
  Wang, Lu~Wang, and Weizhu Chen.
\newblock Lora: Low-rank adaptation of large language models.
\newblock In \emph{ICLR}, 2022.

\bibitem[Hu et~al.(2014)Hu, Xu, and Yang]{hu2014joint}
Zhe Hu, Li~Xu, and Ming-Hsuan Yang.
\newblock Joint depth estimation and camera shake removal from single blurry
  image.
\newblock In \emph{CVPR}, 2014.

\bibitem[Im et~al.(2023)Im, Shin, Baik, and Kim]{im2023deep}
Eun~Woo Im, Junseong Shin, Sungyong Baik, and Tae~Hyun Kim.
\newblock Deep variational bayesian modeling of haze degradation process.
\newblock In \emph{CIKM}, 2023.

\bibitem[Jin et~al.(2018)Jin, Meishvili, and Favaro]{jin2018learning}
Meiguang Jin, Givi Meishvili, and Paolo Favaro.
\newblock Learning to extract a video sequence from a single motion-blurred
  image.
\newblock In \emph{CVPR}, 2018.

\bibitem[Kingma and Ba(2015)]{kingma2015adam}
Diederik~P. Kingma and Jimmy Ba.
\newblock Adam: A method for stochastic optimization.
\newblock In \emph{ICLR}, 2015.

\bibitem[Kingma and Welling(2014)]{kingma2013auto}
Diederik~P. Kingma and Max Welling.
\newblock Auto-encoding variational bayes.
\newblock In \emph{ICLR}, 2014.

\bibitem[Kong et~al.(2023)Kong, Dong, Ge, Li, and Pan]{kong2023efficient}
Lingshun Kong, Jiangxin Dong, Jianjun Ge, Mingqiang Li, and Jinshan Pan.
\newblock Efficient frequency domain-based transformers for high-quality image
  deblurring.
\newblock In \emph{CVPR}, 2023.

\bibitem[Kupyn et~al.(2019)Kupyn, Martyniuk, Wu, and Wang]{kupyn2019deblurgan}
Orest Kupyn, Tetiana Martyniuk, Junru Wu, and Zhangyang Wang.
\newblock Deblurgan-v2: Deblurring (orders-of-magnitude) faster and better.
\newblock In \emph{ICCV}, 2019.

\bibitem[Lin et~al.(2024)Lin, He, Chen, Lyu, Dai, Yu, Ouyang, Qiao, and
  Dong]{lin2024diffbir}
Xinqi Lin, Jingwen He, Ziyan Chen, Zhaoyang Lyu, Bo~Dai, Fanghua Yu, Wanli
  Ouyang, Yu~Qiao, and Chao Dong.
\newblock Diffbir: Towards blind image restoration with generative diffusion
  prior.
\newblock In \emph{ECCV}, 2024.

\bibitem[Lipman et~al.(2023)Lipman, Chen, Ben-Hamu, Nickel, and
  Le]{lipman2023flow}
Yaron Lipman, Ricky T.~Q. Chen, Heli Ben-Hamu, Maximilian Nickel, and Matthew
  Le.
\newblock Flow matching for generative modeling.
\newblock In \emph{ICLR}, 2023.

\bibitem[Liu et~al.(2026)Liu, Zhou, Xu, Cao, Chen, and Zhang]{liu2025fidediff}
Xiaoyang Liu, Zhengyan Zhou, Zihang Xu, Jiezhang Cao, Zheng Chen, and Yulun
  Zhang.
\newblock Fidediff: Efficient diffusion model for high-fidelity image motion
  deblurring.
\newblock In \emph{ICLR}, 2026.

\bibitem[Liu et~al.(2023)Liu, Gong, and Liu]{liu2023rectified}
Xingchao Liu, Chengyue Gong, and Qiang Liu.
\newblock Flow straight and fast: Learning to generate and transfer data with
  rectified flow.
\newblock In \emph{ICLR}, 2023.

\bibitem[Liu et~al.(2024)Liu, Ke, Liu, Zhao, and Lau]{liu2024diffplugin}
Yuhao Liu, Zhanghan Ke, Fang Liu, Nanxuan Zhao, and Rynson W.~H. Lau.
\newblock Diff-plugin: Revitalizing details for diffusion-based low-level
  tasks.
\newblock In \emph{CVPR}, 2024.

\bibitem[Loshchilov and Hutter(2017)]{loshchilov2017sgdr}
Ilya Loshchilov and Frank Hutter.
\newblock {SGDR}: Stochastic gradient descent with warm restarts.
\newblock In \emph{ICLR}, 2017.

\bibitem[Lu et~al.(2026)Lu, Lu, Sun, Zhao, Jiang, Wang, Li, Geng, and
  He]{lu2026one}
Yiyang Lu, Susie Lu, Qiao Sun, Hanhong Zhao, Zhicheng Jiang, Xianbang Wang,
  Tianhong Li, Zhengyang Geng, and Kaiming He.
\newblock One-step latent-free image generation with pixel mean flows.
\newblock \emph{arXiv preprint arXiv:2601.22158}, 2026.

\bibitem[Ma et~al.(2022)Ma, Li, Liao, Zhang, Wang, Wang, and
  Sander]{ma2022deblurnerf}
Li~Ma, Xiaoyu Li, Jing Liao, Qi~Zhang, Xuan Wang, Jue Wang, and Pedro~V.
  Sander.
\newblock Deblur-{NeRF}: Neural radiance fields from blurry images.
\newblock In \emph{CVPR}, 2022.

\bibitem[Mao et~al.(2024)Mao, Li, and Wang]{mao2024adarevd}
Xintian Mao, Qingli Li, and Yan Wang.
\newblock Adarevd: Adaptive patch exiting reversible decoder pushes the limit
  of image deblurring.
\newblock In \emph{CVPR}, 2024.

\bibitem[Nah et~al.(2017)Nah, Kim, and Lee]{nah2017deep}
Seungjun Nah, Tae~Hyun Kim, and Kyoung~Mu Lee.
\newblock Deep multi-scale convolutional neural network for dynamic scene
  deblurring.
\newblock In \emph{CVPR}, 2017.

\bibitem[Pan et~al.(2016)Pan, Sun, Pfister, and Yang]{pan2016blind}
Jinshan Pan, Deqing Sun, Hanspeter Pfister, and Ming-Hsuan Yang.
\newblock Blind image deblurring using dark channel prior.
\newblock In \emph{CVPR}, 2016.

\bibitem[Pham et~al.(2023)Pham, Tran, Tran, Pham, Nguyen, and
  Hoai]{pham2023hypercut}
Bang-Dang Pham, Phong Tran, Anh Tran, Cuong Pham, Rang Nguyen, and Minh Hoai.
\newblock Hypercut: Video sequence from a single blurry image using
  unsupervised ordering.
\newblock In \emph{CVPR}, 2023.

\bibitem[Purohit et~al.(2019)Purohit, Shah, and
  Rajagopalan]{purohit2019bringing}
Kuldeep Purohit, Anshul Shah, and A.~N. Rajagopalan.
\newblock Bringing alive blurred moments.
\newblock In \emph{CVPR}, 2019.

\bibitem[Rim et~al.(2020)Rim, Lee, Won, and Cho]{rim2020realblur}
Jaesung Rim, Haeyun Lee, Jucheol Won, and Sunghyun Cho.
\newblock Real-world blur dataset for learning and benchmarking deblurring
  algorithms.
\newblock In \emph{ECCV}, 2020.

\bibitem[Rombach et~al.(2022)Rombach, Blattmann, Lorenz, Esser, and
  Ommer]{rombach2022high}
Robin Rombach, Andreas Blattmann, Dominik Lorenz, Patrick Esser, and Bj{\"o}rn
  Ommer.
\newblock High-resolution image synthesis with latent diffusion models.
\newblock In \emph{CVPR}, 2022.

\bibitem[Sayed and Brostow(2021)]{sayed2021motionblur}
Mohamed Sayed and Gabriel Brostow.
\newblock Improved handling of motion blur in online object detection.
\newblock In \emph{CVPR}, 2021.

\bibitem[Shen et~al.(2019)Shen, Wang, Lu, Shen, Ling, Xu, and
  Shao]{shen2019hide}
Ziyi Shen, Wenguan Wang, Xiankai Lu, Jianbing Shen, Haibin Ling, Tingfa Xu, and
  Ling Shao.
\newblock Human-aware motion deblurring.
\newblock In \emph{ICCV}, 2019.

\bibitem[Shin and Kim(2026)]{shin2026prism}
Junseong Shin and Tae~Hyun Kim.
\newblock {PRISM}: Latent composition consistency for single-image reflection
  removal.
\newblock In \emph{ECCV}, 2026.

\bibitem[Shin et~al.(2025)Shin, Chung, Yang, and Kim]{shin2025hazeflow}
Junseong Shin, Seungwoo Chung, Yunjeong Yang, and Tae~Hyun Kim.
\newblock {HazeFlow}: Revisit haze physical model as ode and non-homogeneous
  haze generation for real-world dehazing.
\newblock In \emph{ICCV}, 2025.

\bibitem[Shin et~al.(2026)Shin, Kim, Kim, Kim, and Kim]{shin2026oddr}
Junseong Shin, Kijun Kim, Minseong Kim, Dongjin Kim, and Tae~Hyun Kim.
\newblock {ODDR}: One-step deshadow diffusion via reward guidance.
\newblock In \emph{NeurIPS}, 2026.

\bibitem[Sim et~al.(2021)Sim, Oh, and Kim]{sim2021xvfi}
Hyeonjun Sim, Jihyong Oh, and Munchurl Kim.
\newblock {XVFI}: extreme video frame interpolation.
\newblock In \emph{ICCV}, 2021.

\bibitem[Tao et~al.(2018)Tao, Gao, Wang, Shen, Wang, and Jia]{tao2018scale}
Xin Tao, Hongyun Gao, Yi~Wang, Xiaoyong Shen, Jue Wang, and Jiaya Jia.
\newblock Scale-recurrent network for deep image deblurring.
\newblock In \emph{CVPR}, 2018.

\bibitem[Tedla et~al.(2025)Tedla, Zhu, Canham, Taubner, Brown, Kutulakos, and
  Lindell]{tedla2025blur2vid}
SaiKiran Tedla, Kelly Zhu, Trevor Canham, Felix Taubner, Michael~S. Brown,
  Kiriakos~N. Kutulakos, and David~B. Lindell.
\newblock Generating the past, present and future from a motion-blurred image.
\newblock \emph{ACM Transactions on Graphics (Blur2Vid)}, 44\penalty0 (6),
  2025.

\bibitem[Teed and Deng(2020)]{teed2020raft}
Zachary Teed and Jia Deng.
\newblock {RAFT}: Recurrent all-pairs field transforms for optical flow.
\newblock In \emph{ECCV}, 2020.

\bibitem[Tsai et~al.(2022)Tsai, Peng, Lin, Tsai, and Lin]{tsai2022stripformer}
Fu-Jen Tsai, Yan-Tsung Peng, Yen-Yu Lin, Chung-Chi Tsai, and Chia-Wen Lin.
\newblock Stripformer: Strip transformer for fast image deblurring.
\newblock In \emph{ECCV}, 2022.

\bibitem[Unterthiner et~al.(2019)Unterthiner, van Steenkiste, Kurach, Marinier,
  Michalski, and Gelly]{unterthiner2019fvd}
Thomas Unterthiner, Sjoerd van Steenkiste, Karol Kurach, Raphael Marinier,
  Marcin Michalski, and Sylvain Gelly.
\newblock {FVD}: A new metric for video generation.
\newblock \emph{ICLR Workshop}, 2019.

\bibitem[Vaswani et~al.(2017)Vaswani, Shazeer, Parmar, Uszkoreit, Jones, Gomez,
  Kaiser, and Polosukhin]{vaswani2017attention}
Ashish Vaswani, Noam Shazeer, Niki Parmar, Jakob Uszkoreit, Llion Jones,
  Aidan~N. Gomez, Lukasz Kaiser, and Illia Polosukhin.
\newblock Attention is all you need.
\newblock In \emph{NeurIPS}, 2017.

\bibitem[Wang et~al.(2004)Wang, Bovik, Sheikh, and Simoncelli]{wang2004ssim}
Zhou Wang, Alan~C Bovik, Hamid~R Sheikh, and Eero~P Simoncelli.
\newblock Image quality assessment: from error visibility to structural
  similarity.
\newblock \emph{IEEE Transactions on Image Processing}, 13\penalty0
  (4):\penalty0 600--612, 2004.

\bibitem[Wu et~al.(2024)Wu, Sun, Ma, and Zhang]{wu2024osediff}
Rongyuan Wu, Lingchen Sun, Zhiyuan Ma, and Lei Zhang.
\newblock One-step effective diffusion network for real-world image
  super-resolution.
\newblock In \emph{NeurIPS}, 2024.

\bibitem[Zamir et~al.(2021)Zamir, Arora, Khan, Hayat, Khan, Yang, and
  Shao]{zamir2021mprnet}
Syed~Waqas Zamir, Aditya Arora, Salman Khan, Munawar Hayat, Fahad~Shahbaz Khan,
  Ming-Hsuan Yang, and Ling Shao.
\newblock Multi-stage progressive image restoration.
\newblock In \emph{CVPR}, 2021.

\bibitem[Zamir et~al.(2022)Zamir, Arora, Khan, Hayat, Khan, and
  Yang]{zamir2022restormer}
Syed~Waqas Zamir, Aditya Arora, Salman Khan, Munawar Hayat, Fahad~Shahbaz Khan,
  and Ming-Hsuan Yang.
\newblock Restormer: Efficient transformer for high-resolution image
  restoration.
\newblock In \emph{CVPR}, 2022.

\bibitem[Zhang et~al.(2019)Zhang, Dai, Li, and Koniusz]{zhang2019deep}
Hongguang Zhang, Yuchao Dai, Hongdong Li, and Piotr Koniusz.
\newblock Deep stacked hierarchical multi-patch network for image deblurring.
\newblock In \emph{CVPR}, 2019.

\bibitem[Zhang et~al.(2018)Zhang, Isola, Efros, Shechtman, and
  Wang]{zhang2018lpips}
Richard Zhang, Phillip Isola, Alexei~A. Efros, Eli Shechtman, and Oliver Wang.
\newblock The unreasonable effectiveness of deep features as a perceptual
  metric.
\newblock In \emph{CVPR}, 2018.

\bibitem[Zhang et~al.(2021)Zhang, Wang, Maybank, and Tao]{zhang2021motionetr}
Youjian Zhang, Chaoyue Wang, Stephen~J. Maybank, and Dacheng Tao.
\newblock Exposure trajectory recovery from motion blur.
\newblock \emph{IEEE Transactions on Pattern Analysis and Machine
  Intelligence}, 2021.

\bibitem[Zhong et~al.(2022)Zhong, Sun, Wu, Zheng, Lin, and
  Sato]{zhong2022animation}
Zhihang Zhong, Xiao Sun, Zhirong Wu, Yinqiang Zheng, Stephen Lin, and Imari
  Sato.
\newblock Animation from blur: Multi-modal blur decomposition with motion
  guidance.
\newblock In \emph{ECCV}, 2022.

\end{thebibliography}

\newpage
\appendix

\section{Training Details}
\label{app:training}

\paragraph{Architecture and trainable parameters.}
We instantiate $f_\theta$ on FLUX.2-klein-4B~\cite{flux2024}, a $\sim$4B-parameter diffusion transformer paired with a pretrained VAE $(\mathcal{E}, \mathcal{D})$.
We perform full fine-tuning of the transformer only (no LoRA~\cite{hu2022lora} or partial freezing); the VAE encoder and decoder are kept frozen throughout training.
$f_\theta$ is parameterized as a residual displacement from the encoded blur input,
$f_\theta(z_\text{blur}, s, t) := z_\text{blur} + h_\theta(z_\text{blur}, s, t),$
where $h_\theta$ is the transformer output.
Interval conditioning replaces FLUX's original timestep conditioning: the triple $(s, t, t-s)$ is sinusoidally embedded, projected by a shared MLP, and injected through the existing AdaLN pathway.

\paragraph{Data pipeline.}
We use the GoPro training split of 2{,}103 blurry/sharp pairs, where each blurry image is paired with 7--13 consecutive subframes that compose it.
At each training step we apply a random crop (with the same crop coordinates shared across the blurry image and all of its paired subframes to preserve pixel correspondence) and a random horizontal flip; no color jitter or other augmentations are used.
Pixel values are normalized to $[0, 1]$ before VAE encoding, matching the input convention used by FLUX.2.
The DataLoader uses 4 workers, and the random seed is fixed for reproducibility.

\paragraph{Loss sampling.}
All three loss terms (Eqs.~\ref{eq:loss_recon}, \ref{eq:loss_add}, \ref{eq:loss_sharp}) are computed in the same batch and averaged with mean reduction:
\begin{itemize}
    \item $\mathcal{L}_\text{recon}$: $(s, t)$ is sampled uniformly from the set of
    subframe-aligned pairs $\{(\tau_i, \tau_j) : 0 \leq i \leq j \leq M-1\}$. With probability
    $0.25$, we constrain to zero-interval pairs $i = j$ (so $s = t = \tau_i$ and
    $\hat{B}(s, t) = S(\tau_i)$), ensuring that the inference regime $s \approx t$
    used for deblurring and pick-a-moment recovery is densely covered during training. While
    inference uses a small but nonzero $\epsilon$ (e.g., $\epsilon = 0.01$), the $s = t$ branch
    and $\mathcal{L}_\text{sharp}$ at $\tau$ both regress to the same sharp-frame target
    $S(\tau)$, so the predictor at $|t - s| \approx \epsilon$ inherits this supervision by
    continuity.
    \item $\mathcal{L}_\text{add}$: the splitting point $m$ is sampled uniformly from the
    subframe timestamps strictly between $s$ and $t$, $\{\tau_k : s < \tau_k < t\}$.
    \item $\mathcal{L}_\text{sharp}$: since ground-truth sharp frames are only available at subframe timestamps, $\tau$ is sampled uniformly from the discrete set $\{i/(M-1) : i = 0, \ldots, M-1\}$, where $M$ is the number of subframes paired with the blurry input. The EA-LPIPS perceptual loss reproduces the implementation of FideDiff~\cite{liu2025fidediff}.
\end{itemize}

\paragraph{Optimization and progressive learning.}
We optimize with Adam~\cite{kingma2015adam} for a total of 100K iterations.
The learning rate is linearly warmed up to $10^{-4}$ over the first 2K steps and then cosine-annealed~\cite{loshchilov2017sgdr} to $10^{-5}$.
We use bf16 mixed precision with gradient accumulation of 2 steps.
Following Restormer~\cite{zamir2022restormer}, we adopt progressive learning, updating (patch size, batch size) pairs to $[(256^2, 8), (384^2, 4), (512^2, 2)]$ at iterations $[30\text{K}, 60\text{K}, 100\text{K}]$.
The auxiliary loss weights $(\lambda_\text{add}, \lambda_\text{sharp}, \lambda_\text{perc}) = (0.1, 1.0, 2.0)$ are linearly warmed up from zero to their target values over the first patch stage (30K iterations), allowing $\mathcal{L}_\text{recon}$ to establish a stable baseline before the structural and boundary constraints take effect.
We do not use Exponential Moving Average (EMA).

\paragraph{Hardware and wall-clock time.}
All training runs on a single NVIDIA RTX A6000 GPU and takes approximately 4 days to complete.

\begin{table}[h]
\centering
\caption{\textbf{Training hyperparameters.}}
\label{tab:hyperparams}
\small
\begin{tabular}{ll}
\toprule
Backbone & FLUX.2-klein-4B (transformer only fine-tuned, VAE frozen) \\
Trainable params & $\sim$4B (full transformer) \\
Optimizer & Adam \\
Total iterations & 100K \\
LR schedule & Linear warmup to $10^{-4}$ over 2K, then cosine-annealing to $10^{-5}$ \\
Mixed precision & bf16 \\
Gradient accumulation & 2 steps \\
Progressive learning & $(256^2, 8) \rightarrow (384^2, 4) \rightarrow (512^2, 2)$ at $30\text{K}/60\text{K}/100\text{K}$ \\
$(\lambda_\text{add}, \lambda_\text{sharp}, \lambda_\text{perc})$ & $(0.1, 1.0, 2.0)$, linear warmup from 0 over first 30K iterations \\
EMA & None \\
Augmentation & Random crop (shared across paired subframes) + horizontal flip \\
Pixel range & $[0, 1]$ before VAE \\
DataLoader workers & 4 \\
Hardware & 1 $\times$ NVIDIA RTX A6000 \\
Training time & $\sim$4 days \\
\bottomrule
\end{tabular}
\end{table}

\section{Empirical Validation of Latent-Space Additivity}
\label{app:additivity_validity}

The additivity loss $\mathcal{L}_\text{add}$ (Eq.~\ref{eq:loss_add}) enforces the identity
$z_{s,t} \approx \omega \, z_{s,m} + (1-\omega)\,z_{m,t}$ in the VAE latent space, where $\omega = (m-s)/(t-s)$ and $z_{x,y} = \mathcal{E}(B(x, y))$.
While the identity holds exactly in pixel space (by Eq.~\ref{eq:blur_additivity}), its validity in latent space depends on the encoder being approximately linear over the manifold of blurry images.
We empirically verify this on the FLUX.2-klein-4B VAE that we use throughout the paper.
Approximate compositionality in a pretrained VAE latent space has also been exploited for reflection removal~\cite{shin2026prism}.

\paragraph{Protocol.}
We use the GoPro test split's high-frame-rate subframes ($11$ sequences, $\sim$1{,}100 frames each).
For each sequence we sample $30$ random $11$-frame windows, with a shared $512\!\times\!512$ random crop per window.
For each window we sample $10$ random triples $(s, t, m)$ with $0 \le s < m < t \le 1$, $|t-s| \ge 0.2$, and $|m - s|, |t - m| \ge 0.05$, yielding $2{,}983$ triples in total.
We deliberately exclude very small intervals from this protocol because the additivity identity governs the regime where multiple subframes are averaged into each side of Eq.~\ref{eq:blur_additivity}; the zero-interval limit $s \approx t$ is supervised separately by $\mathcal{L}_\text{sharp}$ (and by the $s = t$ branch of $\mathcal{L}_\text{recon}$) directly against the ground-truth sharp frame, so it is not governed by the additivity assumption being validated here.
For each triple we compute $B_{s,t}$, $B_{s,m}$, $B_{m,t}$ as the average of the subframes whose normalized timestamp falls in the corresponding interval, encode each through the frozen VAE to obtain $z_{s,t}$, $z_{s,m}$, $z_{m,t}$, and then form the additive prediction $z_\text{pred} = \omega z_{s,m} + (1-\omega)\,z_{m,t}$.
We decode both $z_{s,t}$ and $z_\text{pred}$ back to pixel space and report PSNR against the corresponding pixel-space references.

\paragraph{Metrics.}
Three pixel-space PSNR quantities isolate the source of error:
\begin{itemize}
    \item $\textsf{PSNR}\bigl(\mathcal{D}(z_{s,t}), B_{s,t}\bigr)$ measures the VAE round-trip error alone (the upper bound achievable without any additivity assumption);
    \item $\textsf{PSNR}\bigl(\mathcal{D}(z_\text{pred}), \mathcal{D}(z_{s,t})\bigr)$ isolates the pure additivity error in pixel space, factoring out the round-trip degradation;
    \item $\textsf{PSNR}\bigl(\mathcal{D}(z_\text{pred}), B_{s,t}\bigr)$ measures the combined error a downstream consumer would observe.
\end{itemize}

\paragraph{Results.}
Table~\ref{tab:additivity_validity} summarizes the three PSNR distributions over $2{,}983$ triples.
The VAE round-trip alone yields a mean PSNR of $42.70$\,dB, while the additive prediction $\mathcal{D}(z_\text{pred})$ matches the round-trip output at $39.82$\,dB on average. The additivity assumption thus introduces an additional $\sim$3\,dB error on top of the VAE's intrinsic reconstruction floor, roughly twice its MSE.
This is of the same order as the round-trip error itself, so it is not a new dominant source of error, and it enters only as a training signal rather than as a constraint at inference.
A pixel-space variant would avoid this approximation but require additional decoder evaluations at each training step.
The combined error against the true pixel target $B_{s,t}$ remains at $37.40$\,dB, comfortably above the noise floor at which deblurring metrics saturate.
We further verified that the additivity error is essentially uncorrelated with both the interval width $|t-s|$ and the splitting position $\omega$, indicating that the encoder's near-linearity is uniform over the parameter range used during training rather than concentrated near a degenerate corner.
These observations justify applying $\mathcal{L}_\text{add}$ directly in the latent space without an explicit decode step.

\begin{table}[h]
\centering
\caption{\textbf{Latent-space additivity validity.} PSNR distributions over $N = 2{,}983$ random $(s, t, m)$ triples drawn from $11$ GoPro test sequences ($512\!\times\!512$ crops). The pure additivity error $\mathcal{D}(z_\text{pred})$ vs. $\mathcal{D}(z_{s,t})$ is comparable to the VAE round-trip floor, indicating that $\mathcal{E}$ is approximately linear over the blur-image manifold.}
\label{tab:additivity_validity}
\small
\begin{tabular}{l|c|ccccc}
\toprule
PSNR (dB) & Description & Mean & Median & Std. & 10th pct. & 90th pct. \\
\midrule
$\mathcal{D}(z_{s,t})$ vs. $B_{s,t}$ & \quad (VAE round-trip only)              & 42.70 & 43.12 & 2.32 & 39.27 & 45.24 \\
$\mathcal{D}(z_\text{pred})$ vs. $\mathcal{D}(z_{s,t})$ & \quad (pure additivity) & 39.82 & 39.38 & 4.49 & 34.36 & 45.94 \\
$\mathcal{D}(z_\text{pred})$ vs. $B_{s,t}$ & \quad (combined)                     & 37.40 & 37.26 & 2.80 & 33.83 & 41.03 \\
\bottomrule
\end{tabular}
\end{table}

\section{Center Frame Quality on GoPro-7}
\label{app:center_gopro7}
We additionally report the quality of the center sharp frame produced by each blur-to-video method on the GoPro-7 split (1{,}744 clips at $640\times360$ resolution, the same evaluation protocol as Table~\ref{tab:main_seq}; frame index 3 of the 7-frame sequence is taken as the center).
This complements Table~\ref{tab:main_seq}'s sequence-level metrics by isolating per-frame deblurring quality at the exposure midpoint within each method's recovered sequence.
Table~\ref{tab:center_gopro7} reports full-image PSNR, SSIM, LPIPS, and DISTS against the corresponding ground-truth sharp frame.
\emph{Note: this evaluation is not directly comparable to Table~\ref{tab:main_deblur}, which uses the standard GoPro test split with 11-frame averaging at full resolution; the protocols differ in averaging length, resolution, and test composition.}

\begin{table}[h]
\centering
\caption{\textbf{Center frame quality on the GoPro-7 split.} 1{,}744 clips at $640\times360$. Frame index 3 of the recovered 7-frame sequence is evaluated against the corresponding ground-truth sharp frame using full-image metrics.}
\label{tab:center_gopro7}
\begin{tabular}{lcccc}
\toprule
Method & PSNR$\uparrow$ & SSIM$\uparrow$ & LPIPS$\downarrow$ & DISTS$\downarrow$ \\
\midrule
Jin et al.~\cite{jin2018learning}        & 29.44 & 0.9237 & 0.1345 & 0.0982 \\
MotionETR~\cite{zhang2021motionetr}      & 28.60 & 0.9415 & 0.0509 & 0.0495 \\
Blur2Vid~\cite{tedla2025blur2vid}        & 30.66 & 0.9528 & 0.0363 & 0.0380 \\
\midrule
Ours                                     & \textbf{33.22} & \textbf{0.9664} & \textbf{0.0291} & \textbf{0.0298} \\
\bottomrule
\end{tabular}
\end{table}

\section{Additional Ablation Studies}
\label{app:ablation}
This section extends the ablations of Sec.~\ref{sec:exp:ablation} with cross-dataset quantitative results and qualitative visualizations.

\paragraph{Cross-dataset loss ablation.}
Table.~\ref{tab:abl_loss_full} extends the loss-term ablation of Table.~\ref{tab:abl_loss} from GoPro to the full evaluation suite (GoPro, HIDE, RealBlur-J, RealBlur-R), measured on single-image deblurring metrics.
The relative ordering established on GoPro holds across all four datasets: $\mathcal{L}_\text{add}$ alone gives marginal gains over $\mathcal{L}_\text{recon}$, $\mathcal{L}_\text{sharp}$ provides the bulk of the perceptual improvement (LPIPS, DISTS), and the full objective achieves the best PSNR/SSIM on every dataset.
This consistency across both synthetic (GoPro, HIDE) and real (RealBlur-J/R) blur indicates that the contribution of each identity-derived loss generalizes beyond the GoPro training distribution.

\begin{table}[h]
\centering
\caption{Cross-dataset loss-term ablation on single-image deblurring metrics. Extension of Table.~\ref{tab:abl_loss}.}
\label{tab:abl_loss_full}
\resizebox{0.8\textwidth}{!}{%
\begin{tabular}{ll cccc}
\toprule
Dataset & Metric & $\mathcal{L}_\text{recon}$ & $\mathcal{L}_\text{recon}+\mathcal{L}_\text{add}$ & $\mathcal{L}_\text{recon}+\mathcal{L}_\text{sharp}$ & Full \\
\midrule
\multirow{4}{*}{GoPro}
  & PSNR$\uparrow$    & 28.45 & 28.56 & 29.24 & \textbf{29.43} \\
  & SSIM$\uparrow$    & 0.9080 & 0.9096 & 0.9236 & \textbf{0.9253} \\
  & LPIPS$\downarrow$ & 0.1455 & 0.1422 & \textbf{0.0787} & 0.0802 \\
  & DISTS$\downarrow$ & 0.1043 & 0.1020 & \textbf{0.0420} & 0.0429 \\
\midrule
\multirow{4}{*}{HIDE}
  & PSNR$\uparrow$    & 26.73 & 26.79 & 27.49 & \textbf{27.52} \\
  & SSIM$\uparrow$    & 0.8668 & 0.8687 & \textbf{0.8894} & 0.8891 \\
  & LPIPS$\downarrow$ & 0.1691 & 0.1682 & \textbf{0.1015} & 0.1046 \\
  & DISTS$\downarrow$ & 0.1284 & 0.1257 & \textbf{0.0539} & 0.0553 \\
\midrule
\multirow{4}{*}{RealBlur-J}
  & PSNR$\uparrow$    & 28.48 & 28.53 & 28.89 & \textbf{28.94} \\
  & SSIM$\uparrow$    & 0.8502 & 0.8518 & 0.8738 & \textbf{0.8757} \\
  & LPIPS$\downarrow$ & 0.1484 & 0.1525 & \textbf{0.1184} & \textbf{0.1184} \\
  & DISTS$\downarrow$ & 0.1121 & 0.1138 & 0.0847 & \textbf{0.0841} \\
\midrule
\multirow{4}{*}{RealBlur-R}
  & PSNR$\uparrow$    & 35.69 & 35.67 & 36.01 & \textbf{36.08} \\
  & SSIM$\uparrow$    & 0.9461 & 0.9444 & 0.9500 & \textbf{0.9510} \\
  & LPIPS$\downarrow$ & 0.1032 & 0.1082 & \textbf{0.0612} & \textbf{0.0612} \\
  & DISTS$\downarrow$ & 0.1314 & 0.1361 & 0.0847 & \textbf{0.0833} \\
\bottomrule
\end{tabular}%
}
\end{table}

\begin{figure}[t]
\centering
\includegraphics[width=\linewidth]{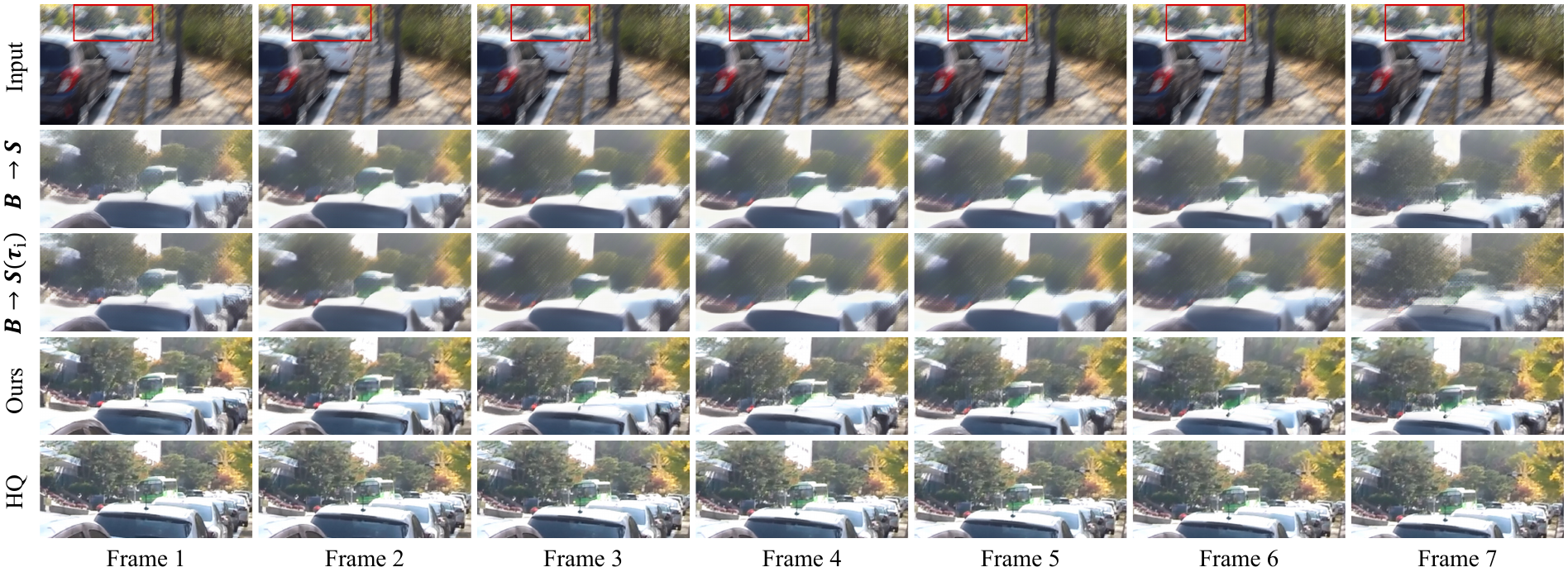}
\caption{Qualitative comparison for the supervision-target ablation on GoPro-7. Best viewed zoomed in.}
\label{fig:abl_inr}
\end{figure}

\paragraph{Advantage of interval prediction.}
Fig.~\ref{fig:abl_inr} provides additional qualitative comparisons for the supervision-target ablation in Table.~\ref{tab:abl_arch}, contrasting the three baselines on a representative GoPro sequence.
The $\mathbf{B \rightarrow S}$ and $(\mathbf{B \rightarrow \{S(\tau_i)\}_{i=0}^{N-1}})$ baselines recover blurrier sequences with weaker per-frame texture, while interval-mean supervision yields sharper content with motion that better aligns with the ground-truth subframes, supporting the quantitative gains in Table.~\ref{tab:abl_arch}.

\begin{figure}[t]
\centering
\includegraphics[width=\linewidth]{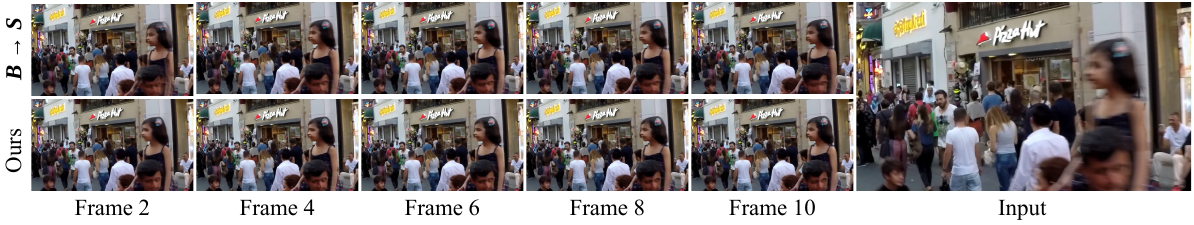}
\caption{Additional blur-to-video comparisons on GoPro, trained on odd frames only.}
\label{fig:abl_odd}
\end{figure}

\paragraph{Generalization to unseen timestamps.}
Fig.~\ref{fig:abl_odd} provides additional qualitative results for the unseen-timestamp ablation in Table.~\ref{tab:abl_time}, where both PickMoment and the $\mathbf{B \rightarrow S}$ baseline are trained only on odd-indexed subframes and evaluated at the held-out even-indexed timestamps.
PickMoment recovers visibly sharper content at the held-out timestamps, indicating that interval-mean supervision generalizes more smoothly across continuous time than direct sharp-frame prediction.

\begin{figure}[t]
\centering
\includegraphics[width=\linewidth]{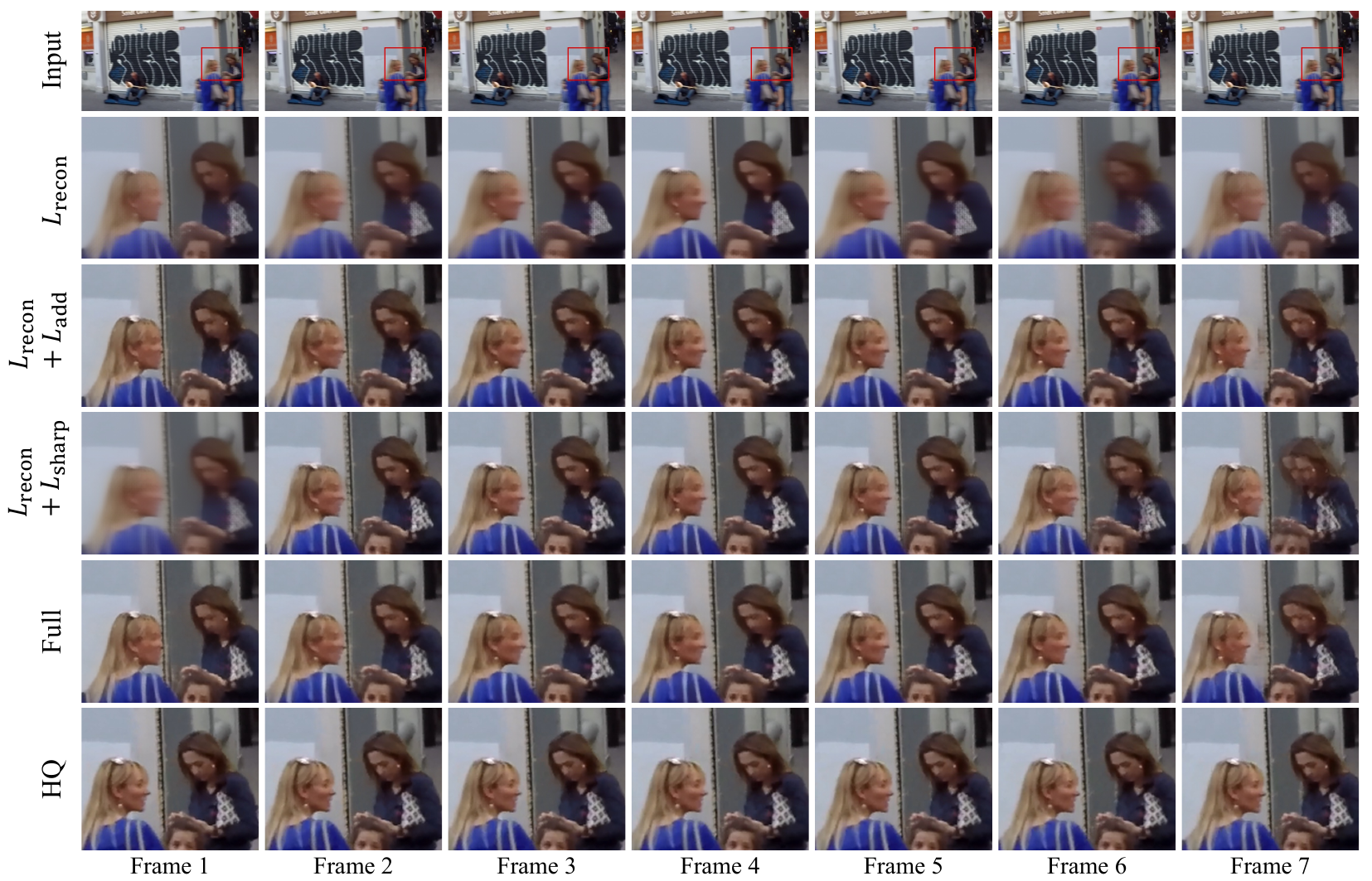}
\caption{Qualitative comparison for the loss-term ablation on GoPro-7 (sequence view). Each row shows a recovered $7$-frame sequence under a different combination of training losses. Best viewed zoomed in.}
\label{fig:abl_loss}
\end{figure}

\begin{figure}[t]
\centering
\includegraphics[width=\linewidth]{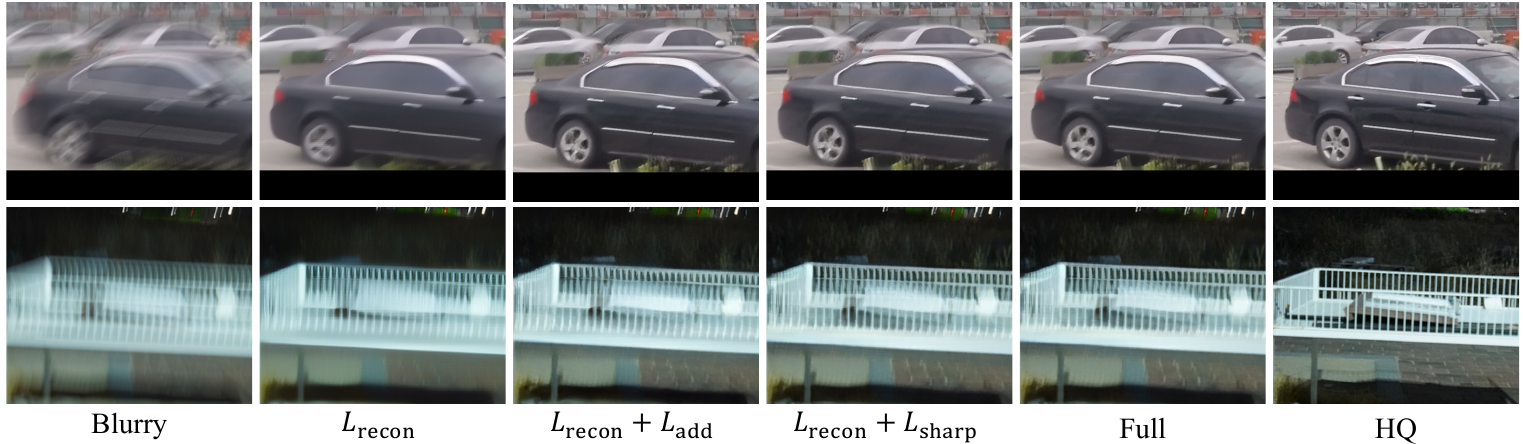}
\caption{Qualitative comparison for the loss-term ablation on GoPro (center-frame deblurring view). Each column shows the recovered center sharp frame under a different combination of training losses. Best viewed zoomed in.}
\label{fig:abl_loss_deblur}
\end{figure}

\paragraph{Effect of each loss term.}
Figs.~\ref{fig:abl_loss} and~\ref{fig:abl_loss_deblur} visualize the complementary roles of $\mathcal{L}_\text{add}$ and $\mathcal{L}_\text{sharp}$ in our full objective (cf.\ Table.~\ref{tab:abl_loss}), from two perspectives.
Fig.~\ref{fig:abl_loss} shows the recovered $7$-frame blur-to-video sequence and reveals the temporal-coherence axis, while Fig.~\ref{fig:abl_loss_deblur} shows the center-frame deblurring output and reveals the per-frame sharpness axis.
Adding $\mathcal{L}_\text{add}$ on top of $\mathcal{L}_\text{recon}$ enforces direction-consistent motion across frames but leaves the center frame blurry, since no loss yet anchors per-frame sharpness.
Adding $\mathcal{L}_\text{sharp}$ instead yields a sharp center frame but with visible inter-frame jitter in the recovered sequence, since each frame is supervised independently with no cross-frame consistency.
Only the full objective combining both losses yields outputs that are sharp \emph{and} direction-consistent simultaneously, demonstrating that the two identity-derived losses address orthogonal failure modes.

\section{Continuous-Time Evidence}
\label{app:continuous_time}

\paragraph{Denser timestamps on unseen data.}
GoPro provides ground truth only at its 7 to 13 subframe timestamps.
We therefore use X4K1000FPS~\cite{sim2021xvfi}: averaging 33 consecutive frames of the 1000-fps source synthesizes the blur and leaves ground truth at 33 timestamps per exposure, over the 15-clip X-TEST split (495 frames per model).
Neither model has seen this data; both are applied zero-shot from their GoPro-trained weights, using the released Blur2Vid checkpoint.
PickMoment outperforms Blur2Vid at every one of the 33 timestamps (Table~\ref{tab:x4k}).

\begin{table}[h]
\centering
\caption{\textbf{Zero-shot PSNR on X4K1000FPS (X-TEST).} Nine of the 33 timestamps per exposure are shown; the last column is the mean over all 33.}
\label{tab:x4k}
\resizebox{\textwidth}{!}{%
\begin{tabular}{l ccccccccc c}
\toprule
$t/T$ & 0.000 & 0.125 & 0.250 & 0.375 & 0.500 & 0.625 & 0.750 & 0.875 & 1.000 & 33-pt mean \\
\midrule
Blur2Vid~\cite{tedla2025blur2vid} & 21.31 & 20.30 & 19.99 & 22.92 & 21.46 & 22.47 & 21.95 & 21.53 & 21.30 & 21.60 \\
Ours & \textbf{22.09} & \textbf{22.39} & \textbf{23.29} & \textbf{23.65} & \textbf{24.07} & \textbf{23.61} & \textbf{22.89} & \textbf{21.82} & \textbf{21.60} & \textbf{22.95} \\
\bottomrule
\end{tabular}%
}
\end{table}

\paragraph{Temporal super-resolution.}
If the model represents a continuous interval-mean field, two queries that name the same physical instant should return the same image even when the input exposure differs.
During training the exposure length of the input blur never changes; only the query interval does.
On the GoPro test set, we synthesize half-exposure blurs by averaging the first half of the subframes.
Route A queries the mid-exposure of the full blur, which is standard center-frame deblurring under the protocol of Table~\ref{tab:main_deblur}; Route B queries the endpoint of the half blur.
Both name the same instant, but the model sees an input length it was never trained on in Route B.

\begin{table}[h]
\centering
\caption{\textbf{Temporal super-resolution on GoPro.} PSNR of two routes that name the same instant, and their mutual agreement.}
\label{tab:temporal_sr}
\begin{tabular}{lc}
\toprule
 & PSNR \\
\midrule
Route A (full blur, queried at mid-exposure) vs.\ GT & 29.43 \\
Route B (half blur, queried at its endpoint) vs.\ GT & 28.49 \\
Agreement (Route A vs.\ Route B) & 27.91 \\
\bottomrule
\end{tabular}
\end{table}

Route B stays within 1 dB of Route A, and the two routes agree more closely ($27.91$ dB) than independent errors at their measured accuracies would predict ($25.9$ dB).
This agreement could also arise from shared model biases, the shared image prior, or the same scene structure, so we treat it as supporting evidence for a continuous interval-mean field rather than proof.

\section{Temporal Consistency and Direction}
\label{app:temporal}

\paragraph{Warping-based temporal consistency.}
To complement FVD and EPE, we warp each recovered frame onto the next along RAFT~\cite{teed2020raft} flow and measure the photometric residual, on the same GoPro-7 clips as Table~\ref{tab:main_seq}.
The ground-truth sequences give the residual of natural motion; a method closer to it neither flickers (rougher than GT) nor over-smooths (smoother than GT).

\begin{table}[h]
\centering
\caption{\textbf{Warping-based temporal consistency on GoPro-7.} Photometric residual after warping each frame onto the next with RAFT flow.}
\label{tab:warp}
\begin{tabular}{lccc}
\toprule
Sequence & warp L1 & warp PSNR & vs.\ GT \\
\midrule
GT (natural motion) & 0.0096 & 35.23 & -- \\
Jin et al.~\cite{jin2018learning} & 0.0106 & 34.52 & rougher than GT \\
Blur2Vid~\cite{tedla2025blur2vid} & 0.0072 & 37.77 & smoother than GT \\
Ours & 0.0089 & 36.49 & closest to GT \\
\bottomrule
\end{tabular}
\end{table}

Ours is close to the ground-truth sequences, so there is no measurable flicker, while Blur2Vid is smoother than the ground truth by a wide margin, consistent with a prior that samples the whole sequence jointly rather than committing per interval.
The per-transition profile for ours is mildly U-shaped ($0.0080$ at mid-exposure, $0.0094$ and $0.0099$ at the two ends), while that of the ground truth is flat.

\paragraph{Motion direction.}
Being deterministic, PickMoment returns the same sequence for a given input, so genuine ambiguity reduces to whether it picks one direction or blends the two.
We score each recovered sequence against the forward and the reversed ground truth.
Across all 1{,}744 GoPro-7 clips, the median PSNR gap between the two scores is $9.2$ dB, and $93\%$ of clips show a gap above $1$ dB, so the model commits to one direction rather than averaging both.
The temporal order of the subframes fixes the direction during supervision, and $\mathcal{L}_\text{add}$ ties the sub-interval predictions to one consistent trajectory.
We report this as an empirical characterization, not a guarantee.

\section{Inference Cost}
\label{app:inference_cost}
We measure inference wall-clock time on a single NVIDIA RTX A6000 GPU at
$1280\!\times\!640$ resolution.
PickMoment generates a $7$-frame sequence in $\sim$8 seconds (mean $7.90$\,s over $10$ runs
after $3$ warmup runs, covering all $7$ forward passes), while Blur2Vid~\cite{tedla2025blur2vid} requires $\sim$3.5 minutes
per video at the default $50$ DDPM steps, yielding roughly a $25\times$ inference speedup
for PickMoment.
This follows the one-step use of generative priors in recent restoration models~\cite{liu2025fidediff,shin2025hazeflow,shin2026oddr}, which likewise replace iterative sampling with a single network evaluation per output.
  
\section{Additional Visual Results}
\label{app:visuals}
We provide additional qualitative comparisons that complement the main paper's figures.
Where applicable, zoomed-in crops are included to highlight texture preservation and motion-trajectory coherence across frames.

\begin{figure}[t]
\centering
\includegraphics[width=\linewidth]{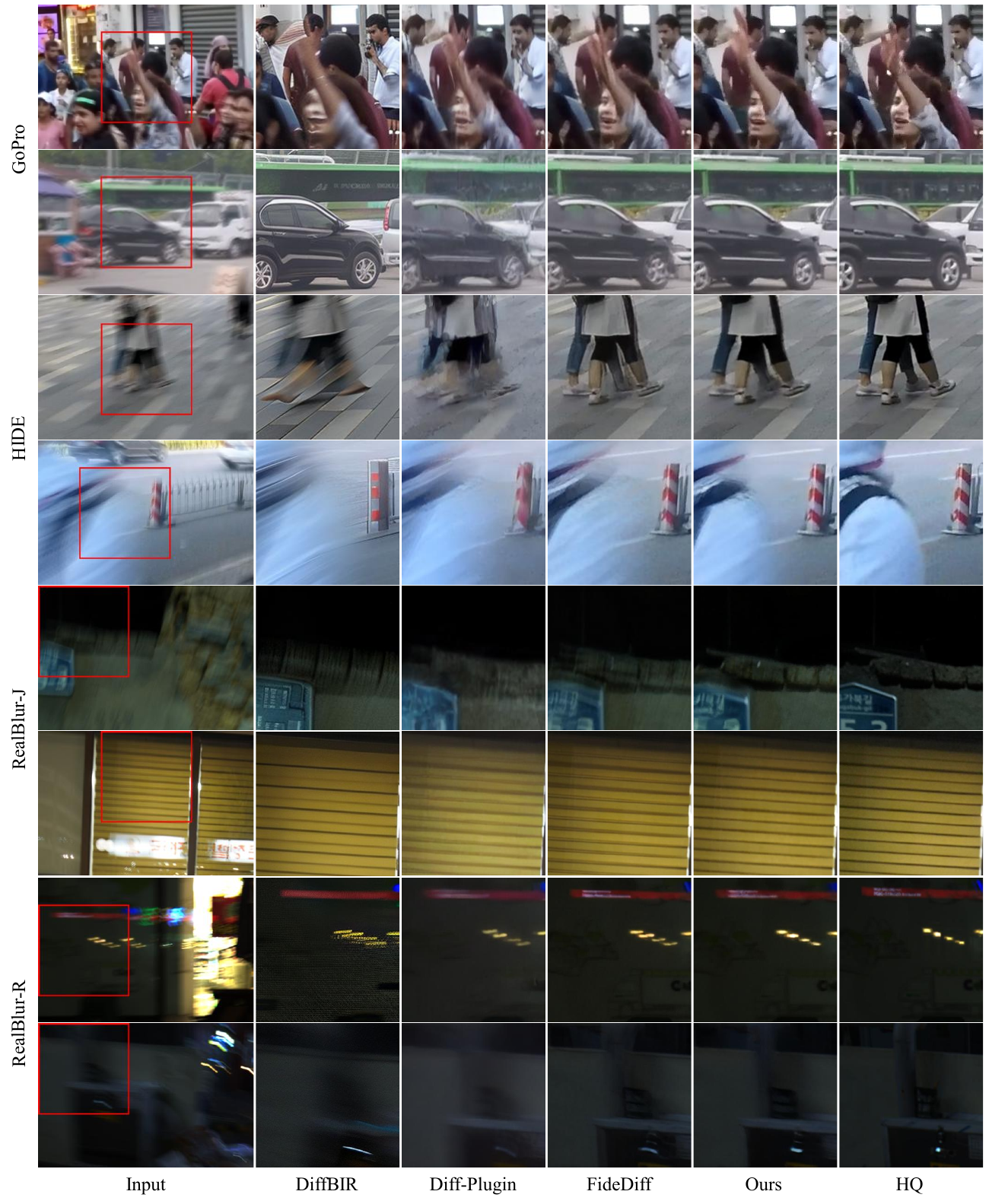}
\caption{Additional single-image deblurring comparisons. Two rows per dataset, GoPro, HIDE, and RealBlur-J, RealBlur-R (top to bottom).}
\label{fig:supp_deblur}
\end{figure}

\paragraph{Single-image deblurring.}
Fig.~\ref{fig:supp_deblur} shows additional single-image deblurring results on GoPro, HIDE, and RealBlur, comparing PickMoment against the strongest baseline from Table.~\ref{tab:main_deblur}.
PickMoment recovers sharper edges and finer texture details than the baseline across all three datasets, including challenging real-blur cases on RealBlur where the blur kernel deviates from the synthetic GoPro distribution.

\begin{figure}[t]
\centering
\includegraphics[width=\linewidth]{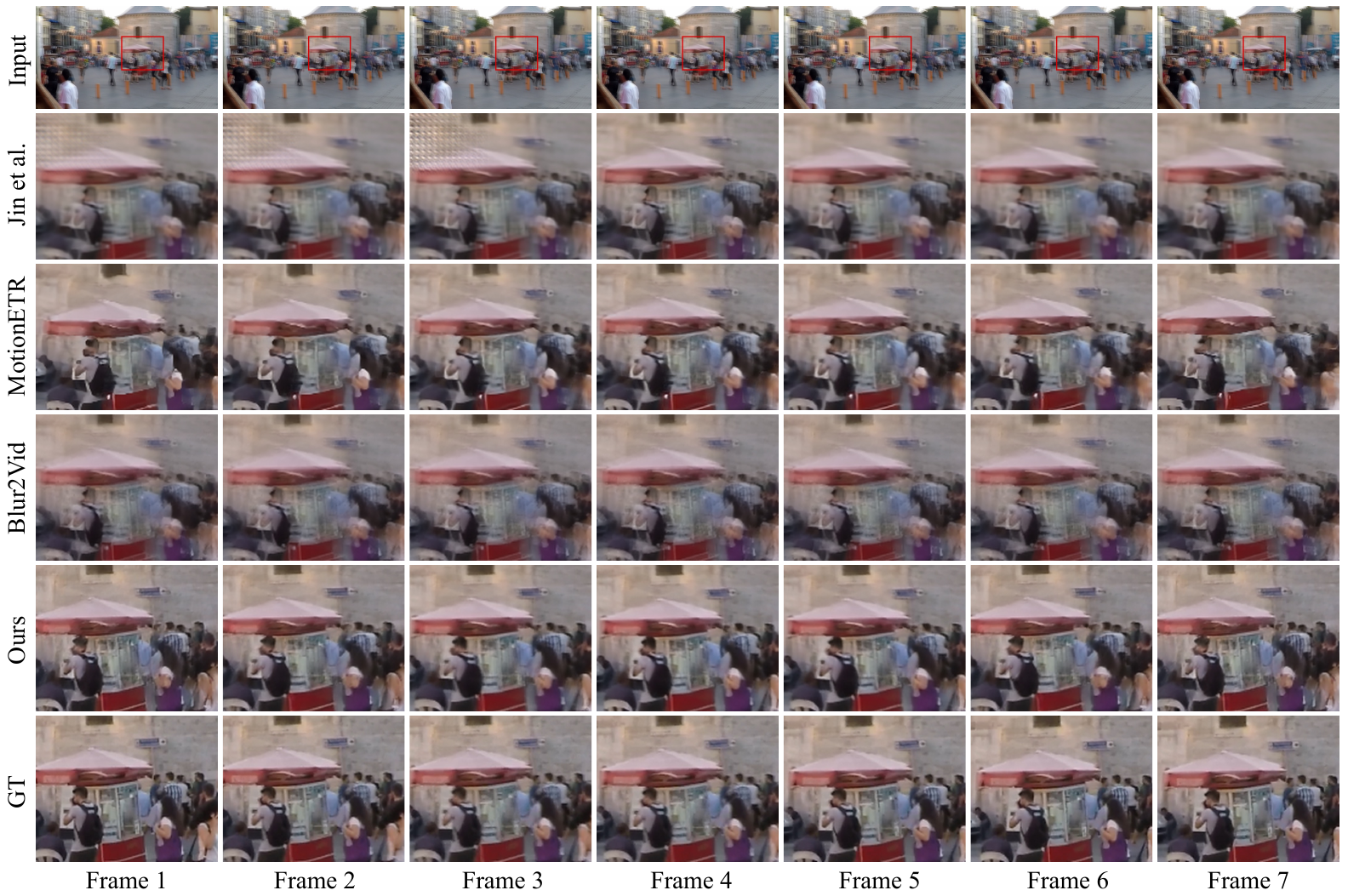}
\caption{Additional blur-to-video comparisons on GoPro-7.}
\label{fig:supp_video1}
\end{figure}

\paragraph{Blur-to-video.}
Fig.~\ref{fig:supp_video1} shows additional blur-to-video sequences on the GoPro-7 split, where each row presents the input blurry image, the recovered $7$-frame sequence by Blur2Vid~\cite{tedla2025blur2vid}, our recovery, and the ground-truth subframes.
PickMoment produces sharper per-frame content with a deterministic, direction-consistent motion trajectory across the recovered sequence.

\section{Limitations}
\label{app:limitations}
We discuss three principal limitations of PickMoment.
\textbf{(i) Single-source supervision.} The model is trained exclusively on GoPro\_Large\_All without auxiliary or domain-augmented data. While it generalizes to HIDE and RealBlur, on RealBlur-J it is competitive but does not consistently outperform FideDiff~\cite{liu2025fidediff}, suggesting headroom from broader training distributions.
\textbf{(ii) Temporal smoothness vs. per-frame fidelity trade-off.} PickMoment predicts each interval independently with one forward pass per interval. While this maximizes per-frame quality (PSNR$_p$, LPIPS$_p$ in Table~\ref{tab:main_seq}), the video-diffusion-based Blur2Vid retains a small advantage on cross-frame smoothness metrics (FVD, EPE), reflecting the implicit temporal regularization of its iterative sampling.
\textbf{(iii) Blur formation assumption.} Eq.~\ref{eq:blur_concept} assumes the blur is well modeled as a uniform temporal integral of sharp frames over a known exposure window, which fits the GoPro synthesis protocol but does not directly cover non-uniform exposures, severe rotational or radial blur, or camera-pipeline effects (e.g., rolling shutter, gamma). Generalization to such regimes is left for future work.
Explicitly modeling the degradation process has proven effective beyond deblurring, e.g., for dehazing~\cite{im2023deep,shin2025hazeflow}.
A natural extension is a weighted interval mean $B_w(s,t) = \int_s^t w(\tau) S(\tau)\,\mathrm{d}\tau \,/ \int_s^t w(\tau)\,\mathrm{d}\tau$ with a calibrated or learned exposure profile $w$, for which the additivity identity still holds with $\omega$ replaced by the ratio of the weight integrals.

\section{Asset Licenses}
\label{app:assets}
Table~\ref{tab:assets} lists the datasets, pretrained models, and software used in this work, along with their licenses. We use each asset in accordance with its license; all assets are properly cited in the main paper.

\begin{table}[h]
\centering
\caption{\textbf{Existing assets used in this work.} Datasets, pretrained models, and software with their licenses.}
\label{tab:assets}
\small
\begin{tabular}{llll}
\toprule
Asset & Type & Citation & License \\
\midrule
GoPro\_Large\_All & Dataset & \cite{nah2017deep} & Research use (per authors) \\
HIDE & Dataset & \cite{shen2019hide} & Research use (per authors) \\
RealBlur-J / RealBlur-R & Dataset & \cite{rim2020realblur} & CC BY-NC-SA 4.0 \\
FLUX.2-klein & Pretrained model & \cite{flux2024} & FLUX.2 Community License (non-commercial) \\
Stable Diffusion (referenced) & Pretrained model & \cite{rombach2022high} & CreativeML Open RAIL-M \\
RAFT & Optical flow & \cite{teed2020raft} & BSD 3-Clause \\
LPIPS & Metric & \cite{zhang2018lpips} & BSD 2-Clause \\
DISTS & Metric & \cite{ding2020dists} & MIT \\
EA-LPIPS & Metric & \cite{liu2025fidediff} & MIT (per authors) \\
PyTorch & Framework & --- & BSD 3-Clause \\
\bottomrule
\end{tabular}
\end{table}

\newpage

\end{document}